\documentclass[journal]{IEEEtran}

\usepackage{xcolor,soul,framed} %,caption

\colorlet{shadecolor}{yellow}
\usepackage[pdftex]{graphicx}
\graphicspath{{../pdf/}{../jpeg/}}
\DeclareGraphicsExtensions{.pdf,.jpeg,.png}

\usepackage[cmex10]{amsmath}
\usepackage{array}
\usepackage{mdwmath}
\usepackage{mdwtab}
\usepackage{eqparbox}
\usepackage{url}
\usepackage{cite}

\usepackage{xspace}
\makeatletter
\DeclareRobustCommand\onedot{\futurelet\@let@token\@onedot}
\def\@onedot{\ifx\@let@token.\else.\null\fi\xspace}

\def\eg{\emph{e.g}\onedot} 
\def\ie{\emph{i.e}\onedot}

\def\etal{\emph{et al}\onedot}
\makeatother

\newcommand \footnoteonlytext[1]
{
    \let \mybackup \thefootnote
    \let \thefootnote \relax
    \footnotetext{#1}
    \let \thefootnote \mybackup
    \let \mybackup \imareallyundefinedcommand
}

\begin{document}
\bstctlcite{IEEEexample:BSTcontrol}
    \title{BANet: A Blur-aware Attention Network for Dynamic Scene Deblurring}
  \author{Fu-Jen~Tsai$^*$,
      Yan-Tsung~Peng$^*$,~\IEEEmembership{Member,~IEEE,}
      Chung-Chi~Tsai,\\
      Yen-Yu~Lin,~\IEEEmembership{Senior Member,~IEEE,}
      Chia-Wen~Lin,~\IEEEmembership{Fellow,~IEEE},\\
      *equal contribution% <-this % stops a space

  \thanks{Manuscript received August 16, 2021; revised May 17, 2022 and August 30, 2022; accepted October 13, 2022. Date of publication month day, 2022; date of current version month day, 2022. This work was funded in part by National Science and Technology Council (NSTC) under grants 110-2634-F-002-050, 111-2628-E-A49-025-MY3, 111-2221-E-004-010, and in part by Qualcomm Technologies, Inc., through a Taiwan University Research Collaboration Project, under Grant NAT-487844. The computational and storage resources for this project were provided in part by National Center for High-performance Computing (NCHC) of National Applied Research Laboratories (NARLabs), Hsinchu, Taiwan. The associate editor coordinating the review of this manuscript and approving it for publication was Dr. Wangmeng Zuo.  (Corresponding author: Chia-Wen Lin)}
  \thanks{F.-J. Tsai is with the Department of Electrical Engineering, National Tsing Hua University, Hsinchu 300044, Taiwan. E-mail: fjtsai@gapp.nthu.edu.tw}
  \thanks{Y.-T. Peng is with the Department of Computer Science, National Chengchi University, Taipei 116011, Taiwan. E-mail: ytpeng@cs.nccu.edu.tw}% <-this % stops a space
  \thanks{C.-C. Tsai is with Qualcomm Technologies, Inc., San Diego, CA 92121, USA. E-mail: chuntsai@qti.qualcomm.com}%
  \thanks{Y.-Y. Lin is with the Department of Computer Science, National Yang Ming Chiao Tung University, Hsinchu 300093, Taiwan. E-mail: lin@cs.nctu.edu.tw}
  \thanks{C.-W. Lin is with the Department of Electrical Engineering, National Tsing Hua University, Hsinchu 300044, Taiwan, and with the Electronic and Optoelectronic System Research Laboratories, Industrial Technology Research Institute, Hsinchu 310401, Taiwan. (e-mail: cwlin@ee.nthu.edu.tw)}% <-this % stops a space
  }

% The paper headers
\markboth{IEEE TRANSACTIONS ON IMAGE PROCESSING}%
{Shell \MakeLowercase{\textit{et al.}}: Bare Demo of IEEEtran.cls for IEEE Journals}

% ====================================================================
\maketitle
\begin{abstract}
Image motion blur results from a combination of object motions and camera shakes, 
and such blurring effect is generally directional and non-uniform.  Previous research attempted to solve non-uniform blurs using self-recurrent multi-scale, multi-patch, or multi-temporal architectures with self-attention to obtain decent results.
However, using self-recurrent frameworks typically leads to a longer inference time, while inter-pixel or inter-channel self-attention may cause excessive memory usage. 
This paper proposes a Blur-aware Attention Network (BANet), that accomplishes accurate and efficient deblurring via a single forward pass. 
Our BANet utilizes region-based self-attention with multi-kernel strip pooling to disentangle blur patterns of different magnitudes and orientations and cascaded parallel dilated convolution to aggregate multi-scale content features. 
Extensive experimental results on the GoPro and RealBlur benchmarks demonstrate that the proposed BANet performs favorably against the state-of-the-arts in blurred image restoration and can provide deblurred results in real-time.
\end{abstract}
%-------------------------------------------------------------------------
\section{Introduction}
\begin{figure}
\begin{center}
\includegraphics[width=1\columnwidth]{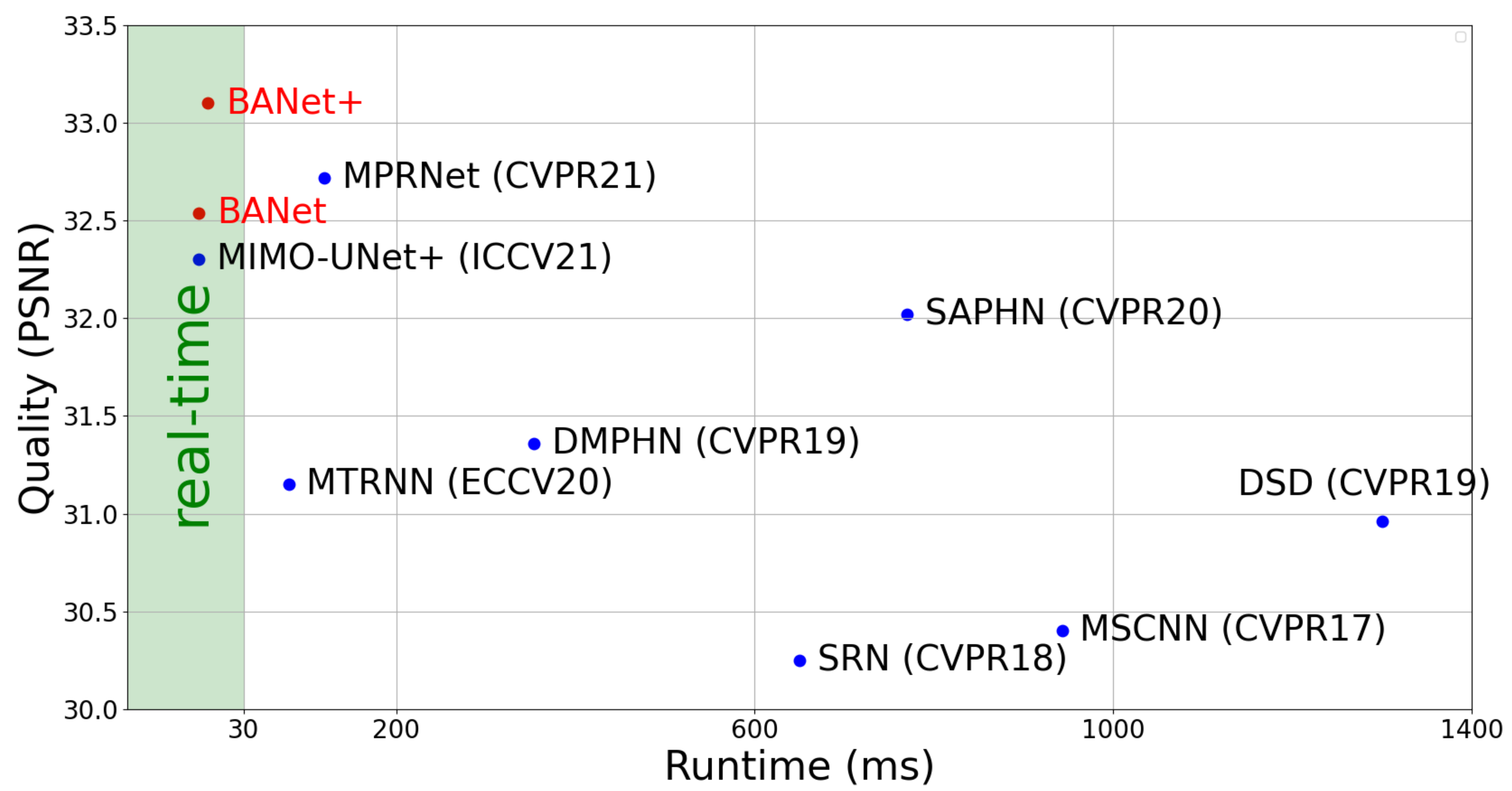}
\end{center}
\caption{Performance comparison on the GoPro test dataset in terms of deblurring quality and runtime complexity. The proposed BANet performs favorably against the state-of-the-art methods in both accuracy and efficiency.}
\label{fig:introduction}
\end{figure}

Dynamic scene deblurring or blind motion deblurring aims to restore a blurred image with little knowledge about the blur kernel. Scene blur caused by camera shakes, object motions, low shutter speeds, or low frame rates not only degrades the quality of taken images/videos but also results in information loss. Therefore, removing such blurring artifacts to recover image details becomes essential to many downstream vision applications, such as facial detection~\cite{Coupled_Learning,Deblurring_Face_Images}, text recognition~\cite{lee2019blind}, moving object segmentation~\cite{pan2019joint}, etc., where clean and sharp images are appreciated. Although significant progress has been made in conventional and deep-learning-based approaches~\cite{Blur_Kernel_Estimation, Nah_2017_CVPR, Zhang_2019_CVPR, MT_2020_ECCV}, we observe a compromise between accuracy and speed. Owing to this observation, we target to develop an efficient and effective algorithm in this paper for blurred image restoration with its current performance in accuracy and speed shown in Fig.~\ref{fig:introduction}.

\begin{figure*}[t!]
\begin{center}
    \includegraphics[width=1\textwidth]{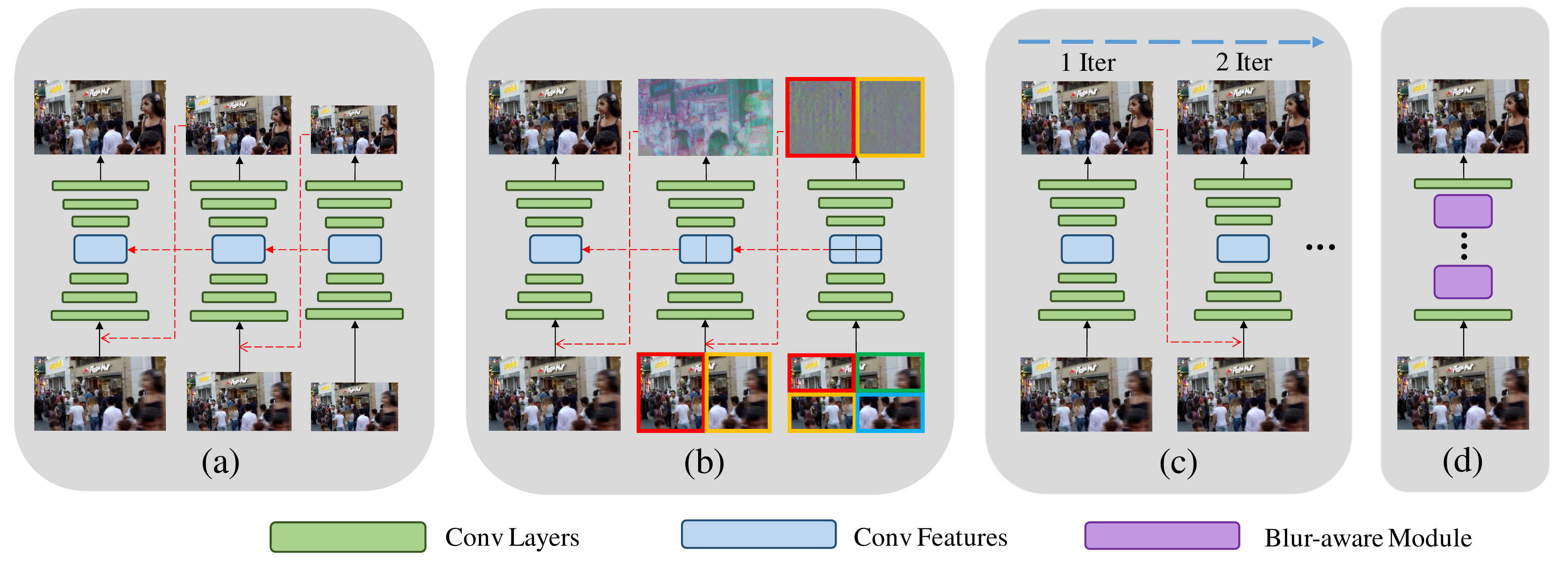}
\end{center}
 \caption{Network architecture comparisons among (a) MS, (b) MP, (c) MT, and (d) our BANet. Recurrent models are typically less efficient. BANet completes deblurring via a single forward pass. }
\label{fig:RNN_type}
\end{figure*} 

Deep-learning-based approaches usually reach superior results, given their better feature representation capability toward dynamic scenes. Among the state-of-the-art architectures for deblurring, self-recurrent models have been widely adopted to leverage blurred image repeatability in either {\em multiple scales}~{(MS)}~\cite{gao2019dynamic, Nah_2017_CVPR, tao2018srndeblur, Dark_and_Bright}, {\em multiple patch levels}~{(MP)}~\cite{Zamir_2021_CVPR,SAPN2020,Zhang_2019_CVPR}, or {\em multiple temporal behaviors}~{(MT)}~\cite{MT_2020_ECCV}, as shown in Fig.~\ref{fig:RNN_type}(a)--(c). Specifically, the {{MS}} models distill multi-scale blur information in a self-recurrent manner and restore blurred images based on the extracted coarse-to-fine features~\cite{gao2019dynamic, Nah_2017_CVPR, tao2018srndeblur}. However, scaling a blurred image to a lower resolution often results in losing edge information~\cite{MT_2020_ECCV}. In contrast, the MP models split a blurred input image into multiple patches to estimate and then remove motion blurs of different scales~\cite{SAPN2020,Zhang_2019_CVPR}.
However, splitting the blurred input and features into equal-sized non-overlapping patches may cause cause contextual information discontinuity, sub-optimal for handling non-uniform blur in dynamic scenes.
%Yet, splitting patches heuristically may cause discontinuities in contextual information. It is sub-optimal for handling non-uniform blur in dynamic scenes.
In~\cite{MT_2020_ECCV}, a self-recurrent MT structure was proposed to progressively eliminate non-uniform blurs over multiple iterations. Each iteration would gradually deblur the image until it becomes sharp. However, its inflexible progressive training and inference process may not generalize well for images of varying region-wise blurring degrees.
Besides, these existing self-recurrent models, including MS, MP, and MT, cannot achieve high-quality deblurring in real-time (say, 30 HD frames per second).

In addition to model architectures, recent research studies~\cite{RADN_2020_ECCV, SAPN2020} further exploit self-attention to address blur non-uniformity. Suin~\etal~\cite{SAPN2020} utilize {MP}-based processing with self-attention to extract features for areas with global and local motions. However, using a self-recurrent mechanism to generate multi-scale features often leads to a significantly longer inference time. To shorten the latency, Purohit and Rajagopalan~\cite{RADN_2020_ECCV} selectively aggregate features through learnable pixel-wise attention~\cite{SAGAN_2019_PMLR} enabled by deformable convolutions for modeling local blurs in a single forward pass. Despite its effectiveness, self-attention exploring pixel-wise or channel-wise correlations via trainable filters often causes high memory usage, thus only applicable to small-scale features~\cite{RADN_2020_ECCV}. Furthermore, motion blurs coming from object motions manifest smeared effects and produce directional and local averaging artifacts, which cannot be handled well by inter-pixel/channel correlations. 

This paper proposes a {\em Blur-aware Attention Network} (BANet) to overcome the above-mentioned issues. BANet is an efficient yet effective single-forward-pass model, as illustrated in Fig.~\ref{fig:RNN_type}(d), which achieves state-of-the-art deblurring performance while working in real-time, as shown in Fig.~\ref{fig:introduction}. Specifically, our model stacks multiple layers of the {\em Blur-Aware Module} (BAM) for removing motion blurs.  
BAM separates the deblurring process into two branches, Blur-aware Attention (BA) and Cascaded Parallel Dilated Convolution (CPDC), where BA locates region-wise blur orientations and magnitudes while CPDC adaptively removes blurs based on the attended blurred features.
Based on an observation of directional and regional averaging artifacts caused by dynamic blurs, the proposed BA derives region-wise attention by using computationally inexpensive regional averaging to capture blurred patterns of different orientations and magnitudes globally and locally. 
To derive the orientations and magnitudes of different blurred regions in an image, we reassemble horizontal and vertical blurred responses to catch irregular blur orientations and utilize multi-scale kernels to learn the magnitudes. CDPC leverages two cascaded multi-scale dilated convolutions to deblur image features.
%It also leverages CPDC to extract features without suffering from blur information loss as an MS model does.
As a result, BANet possesses the superior deblurring capability and can support subsequent real-time applications superbly. 

In short, our contributions are two-fold. First, BANet is featured with a novel BAM module that exploits region-wise attention to capture blur orientations and magnitudes, making BANet capable of disentangling blur contents of different degrees in dynamic scenes. With the disentangled region-wise blurred patterns, it then utilizes cascaded multi-scale dilated convolution to restore blurred features. Second, our efficient single-forward-pass deep networks perform favorably against state-of-the-art methods with fast inference time.
%while running $27$x faster than the visual-quality competitor~\cite{SAPN2020}.
% {\color{red} In short, our contributions are summarized as below:\begin{itemize}
% \item We propose a two branch deblurring module called BAM that can effectively and efficient solving non-uniform blur problems in a single-forward manner.
% \item Observing from the directional and regional averaging artifacts caused by dynamic blurs, we propose a region-wise attention that can capture blurred patterns of different orientations and magnitudes.
% \item Our efficient single-forward-pass deep networks perform favorably against the state-of-the-art methods while running $27$x faster than the best visual-quality competitor~\cite{SAPN2020}.
% \end{itemize}}
\begin{figure*}[t]
    \begin{center}
    \includegraphics[width=0.95\textwidth]{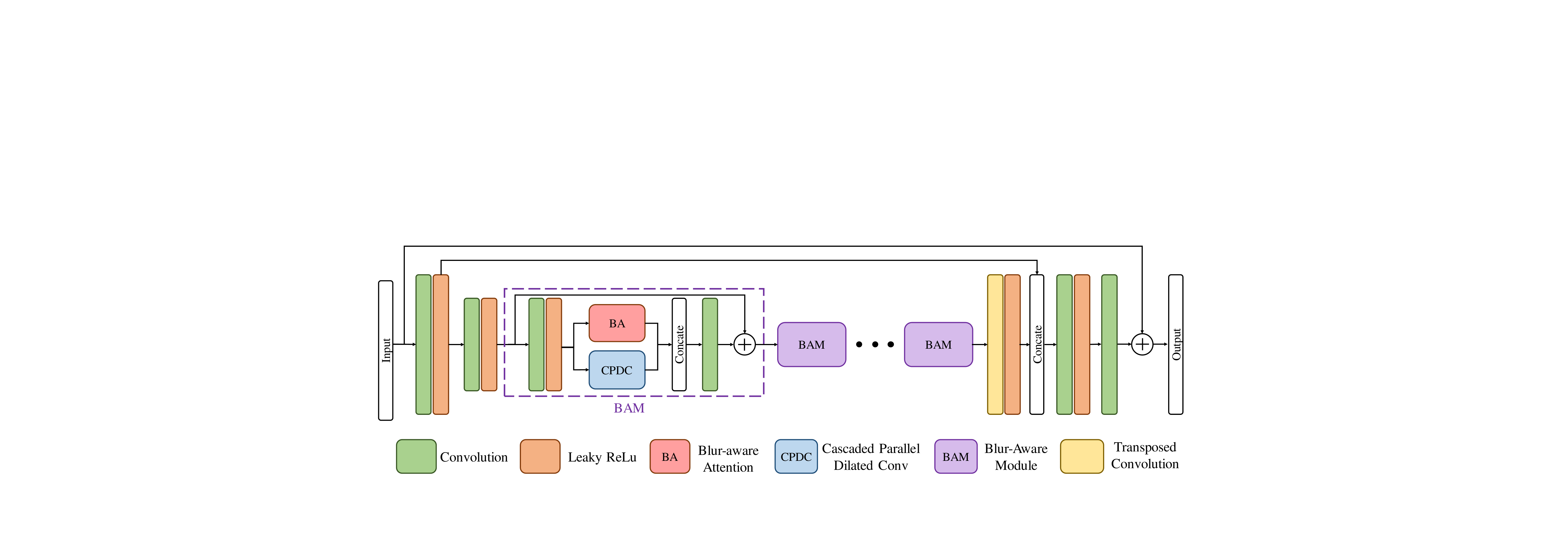}
    \end{center}
    \vspace{-0.1in}
    \caption{Architecture of the proposed blur-aware attention networks (BANet). The blur-aware modules (BAM) serve as the building blocks of BANet. The first BAM is detailed in the purple dotted box while the rest are represented by solid purple boxes. }
    \label{fig:architecture}
\vspace{-0.1in}
\end{figure*}  

\section{Related Work}

\subsection{Conventional Methods}
%\paragraph{Conventional Methods}

    Dynamic Scene image deblurring is a highly ill-posed problem since blurs stem from various factors in the real world. Conventional image deblurring studies often make different assumptions, such as  uniform~\cite{Cho_2009_ACM, Fergus2006, Xu_2010_ECCV, Shan2008} or non-uniform~\cite{gupta_mdf_deblurring, NIPS2010_7f5d04d1, 6126276, 6909750, 5540175} blurs, and image priors~\cite{Chen_2019_CVPR,  5206802, 6909767, Pan_2016_cvpr,8100221}, to model blur characteristics.
    Namely, these methods impose different constraints on the estimated blur kernels, latent images, or both with handcrafted regularization terms for blur removal. Nevertheless, these methods often attempt to solve a non-convex optimization problem and involve heuristic parameter tuning that is entangled with the camera pipeline; thus, they cannot generalize well to complex real-world examples. 
    
\subsection{Deblurring via Learning}
    %charles and yylin
    %\paragraph{Deblurring via Learning}
    
    Learning-based approaches with self-recurrent modules gain great success in single-image deblurring. Particularly, the {\em coarse-to-fine} schemes can gradually restore a sharp image on different resolutions (MS)~\cite{gao2019dynamic, Nah_2017_CVPR, tao2018srndeblur, Dark_and_Bright}, fields of view (MP)~\cite{SAPN2020,Zhang_2019_CVPR}, or temporal characteristics (MT)~\cite{MT_2020_ECCV}. Despite the success, self-recurrent models usually lead to longer inference runtime. Recently, non-recurrent methods~\cite{DeblurGAN, Kupyn_2019_ICCV, RADN_2020_ECCV, Yuan_2020_CVPR, Depth_Guided_Model,MIMO} were proposed for efficient deblurring. For instance, Kupyn~\etal~\cite{DeblurGAN, Kupyn_2019_ICCV} suggested using conditional generative adversarial networks to restore blurred images. However, these methods do not well address non-uniform blurs in dynamic scenes, often causing blur artifacts in the deblurred images. To address this issue, Yuan~\etal~\cite{Yuan_2020_CVPR} proposed a spatially variant deconvolution network with optical flow estimation to guide deformable convolutions and capture moving objects during model training. Li~\etal~\cite{Depth_Guided_Model} proposed a depth-guided model for deblurring.
    However, the optical flow and depth information may not always correlate with blur, which may cause less effective deblurring. Cho~\etal~\cite{MIMO} proposed an efficient multi-scale deblurring structure with a multi-input multi-output. With multi-scale input, the process adopts a shallow convolution to turn the images into attention masks and multiply them by the same scales' features. However, its simple feature attention mechanism may not be able to extract blur information comprehensively from an input image, hence limiting its deblurring performance.

\subsection{Self-attention}
    %charles and yylin
    %\paragraph{Self-attention}

    Self-attention (SA)~\cite{vaswani2017attention} has been widely adopted to advance the fields of image processing~\cite{parmar2018image, SAGAN_2019_PMLR} and computer vision~\cite{hu2018senet, wang2018non}. Recent advances~\cite{RADN_2020_ECCV,SAPN2020} revealed that attention is beneficial for learning inter-pixel correlations to emphasize different local features for removing non-uniform blur. Specifically, Purohit~\etal~\cite{RADN_2020_ECCV} proposed to deblur using SA to explore pixel-wise correlation for non-local feature adaptation. However, since SA requires much memory in $\mathcal{O}(H^2W^2)$ space, where $H$ and $W$ are the height and width of the input to SA, the method can only apply SA to the smallest-scale features (from a $1280\times 720$ blurred input to $160\times 90$ SA's input), limiting the efficacy of SA. Also, motion blurs cause directional and local averaging artifacts, which merely pixel-wise SA may not address well.  Suin~\etal~\cite{SAPN2020} proposed an MP architecture with less memory-intensive SA by using global average pooling with space complexity $\mathcal{O}(d_ad_cHW)$, where $d_a$ is the channel dimension of the components {\em query} and {\em key} in SA, $d_c$ is the dimension of the component {\em value}, and $d_ad_c < HW$.
    Despite the method's less space complexity, compressing pixel information into the channel domain may lose spatial information, thus degrading deblurring performance. In contrast, we propose an efficient and low memory-cost regional averaging SA to capture non-uniform blur information more accurately. It is with space complexity $\mathcal{O}(CHW)$, where $C$ is the number of output channels. It can deblur high-resolution input images and achieve superior performance in real-time.

\section{Proposed Approach}

\begin{figure*}[t!]
\begin{center}
\includegraphics[width=0.95\textwidth]{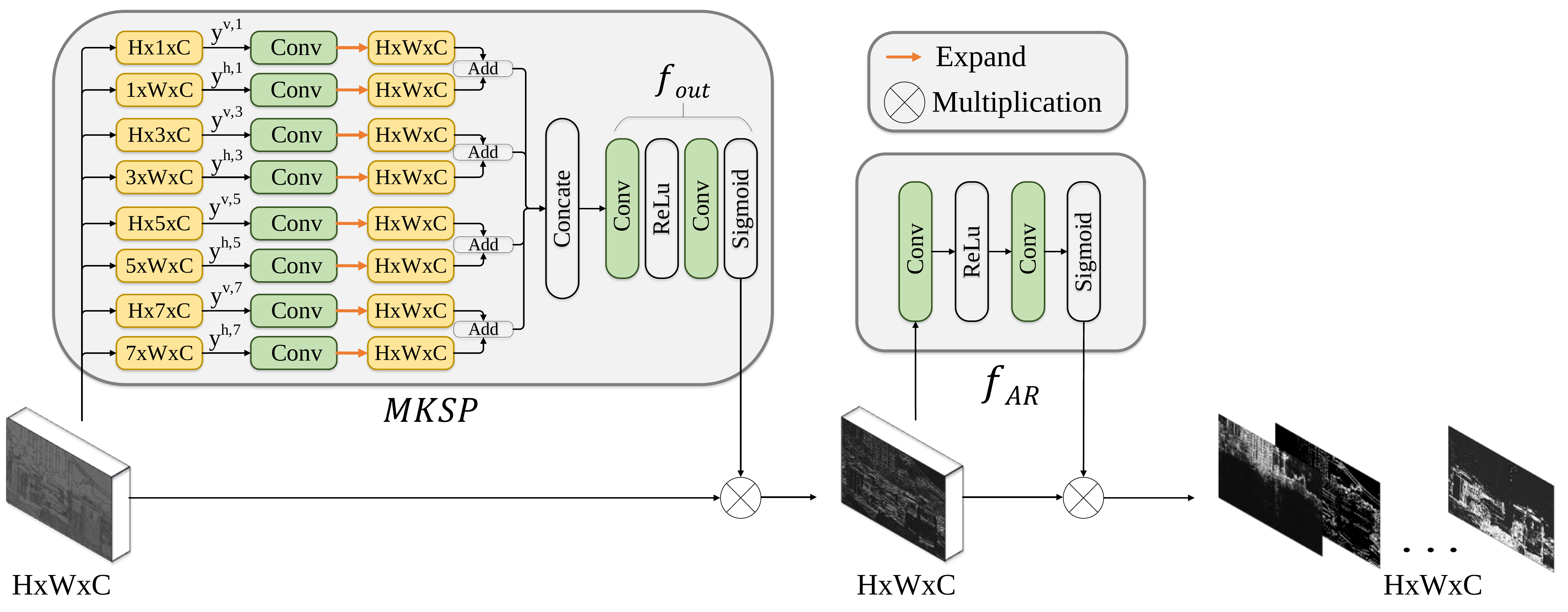}
\vspace{-0.1in}
\end{center}
 \caption{Architecture of blur-aware attention (BA). It cascades two parts, including multi-kernel strip pooling (MKSP) and attention refinement (AR). It is developed to disentangle blurred contents in an efficient way. See the text for details.}
\label{fig:BAA}
\end{figure*}  

We present the blur-aware attention network (BANet) to address the potential issues in two commonly used techniques for deblurring: self-recurrence and self-attention. Self-recurrent algorithms result in longer inference time due to repeatedly accessing input blurred images. Self-attention based on inter-pixel or inter-channel correlations is memory intensive and cannot explicitly capture regional blurring information. Instead, the proposed BANet is a one-pass residual network consisting of a series of stacked blur-aware modules (BAMs), which serve as the building blocks, to disentangle different patterns of blurriness and remove blurs based on the attended blurred features. 
%{\color{red} as the building blocks, which is a two branches architecture, one is for effectively disentangling different patterns of blurriness, the other is for adaptively capturing deblurring information according to the disentangling patterns above.}
%which serve as the building blocks, to effectively disentangle different patterns of blurriness. 

As illustrated in Fig.~\ref{fig:architecture}, BANet starts with two convolutional layers, which contain a stride of $2$ to downsample the input image to half resolution. BANet employs one transposed convolutional layer to upsample features to the original size. In between, we stack a set of BAMs to correlate regions with similar blur and extract multi-scale content features. A BAM consists of two components, BA and CPDC, where BA distills global and local blur orientations and magnitudes, and CPDC captures multi-scale blurred patterns to eliminate blurs adaptively. Combining BA and CPDC, BAM is a residual-like architecture that derives both global and local multi-scale blurring features in a learnable manner. We detail the two key components, BA and CPDC, in the following.

\subsection{Blur-aware Attention (BA)}\label{subsec:BA}

To accurately restore the motion area displaying directional and averaging artifacts caused by object motions and camera shakes, we propose a region-based self-attention module, called BA, to capture such effects in the global (image) and local (patch) scales. As shown in Fig.~\ref{fig:BAA}, BA contains two cascaded parts: multi-kernel strip pooling (MKSP) and attention refinement (AR). MKSP catches multi-scale blurred patterns of different magnitudes and orientations, followed by AR to refine them locally. % We describe their details as follows.

\paragraph{Multi-Kernel Strip Pooling (MKSP)}
Hou \etal~\cite{hou2020strip} presented an SP (strip pooling) method that uses horizontal and vertical one-pixel long kernels to extract long-range band-shape context information for scene parsing. SP averages the input features within a row or a column individually and then fuses the two thin-strip features to discover global cross-region dependencies. 
% Fu-Jen 
%Let the input feature map be a three-dimensional (3D) tensor ${\bf x} = [x_{i,j,c}] \in R^{H\times W\times C}$, 
Let the input feature  maps ${\bf x} = [x_{i,j,c}] \in R^{H\times W\times C}$, where $C$ denotes the number of channels. Applying SP to ${\bf x}$ generates a vertical and a horizontal tensor followed by a 1D convolutional layer with a kernel size of $3$. This produces a vertical tensor ${\bf y}^v = [y_{i,c}^v] \in R^{H\times C}$ and a horizontal tensor ${\bf y}^h = [y_{j,c}^h] \in R^{W\times C}$, where $y_{i,c}^v =   \frac{1}{W}\sum^{W-1}_{j=0} x_{i,j,c}$ and $y_{j,c}^h = \frac{1}{H}\sum^{H-1}_{i=0} x_{i,j,c}$. 
The SP operation, after a convolution layer, fuses the two tensors into ${\bf y} = [y_{i,j,c}] \in R^{H\times W\times C}$, where $y_{i,j,c}=y_{i,c}^v+y_{j,c}^h$, and then turns the fused tensor into an attention mask ${\bf M}_{sp}$ as
\begin{equation}
    {\bf M}_{sp} = \sigma_{sig}(f_{1}({\bf y})),
    \label{eq:SA}
\end{equation}
where $f_1$ is a $1\times 1$ convolutional layer and $\sigma_{sig}(\cdot)$ is the sigmoid function. Although SP has shown its effects on segmenting band-shape objects for scene parsing, it is unsuitable to directly apply SP to an image deblurring task, aiming to locate blurred patterns that tend to involve different orientations and magnitudes, and restore a sharp image.

% Fu-Jen
Motivated by SP, we propose MKSP that adopts strip pooling with different kernel sizes to discover regional and directional averaging artifacts caused by dynamic blurs. 

MKSP combines and compares multiple sizes/scales of averaging results followed by concatenation and convolution to catch blurred patterns of different magnitudes and orientations. The idea behind our design is to reassemble different orientations by horizontal and vertical operations on multi-scale results, \eg, the difference between consecutive kernel sizes, and reveal the scales of blurred patterns. We apply convolutional layers to automatically discover these blur-aware operations on the feature level to learn irregular attended features rather than a fixed cropping method on the image level used in MP methods~\cite{SAPN2020,Zhang_2019_CVPR}. MKSP averages the input tensors within rows and columns by adaptive average pooling to generate $H\times n \times C$ and $n\times W \times C$ long features, where $n \in \{1,3,5,7\}$ represents different scales.
Thus, MKSP generates four pairs of tensors, each of which has a vertical and a horizontal tensor followed by a 1D (for $n=1$) or 2D (for the rest) convolutional layer with the kernel size of $3$ or $3\times 3$, respectively. This produces the vertical tensor ${\bf y}^{v,n}\in R^{H \times n \times C}$ and the horizontal tensor ${\bf y}^{h,n}\in R^{n \times W \times C}$, where the vertical tensor is
\begin{equation}
    y_{i,j,c}^{v,n} =\frac{1}{K_h} \sum_{k=0}^{K_h-1} x_{i, (j\cdot S_h+k),c},
\end{equation}
where the horizontal stride $S_h=\lfloor\frac{W}{n}\rfloor$ and the horizontal-strip kernel size $K_h= W-(n-1)S_h$. Symmetrically, the horizontal tensor is defined by
\begin{equation}
    y_{i,j,c}^{h,n} =\frac{1}{K_v} \sum_{k=0}^{K_v-1} x_{(i\cdot S_v+k), j,c},
\end{equation}
where the vertical stride $S_v = \lfloor\frac{H}{n}\rfloor$ and the vertical-strip kernel size $K_v = H-(n-1)S_v$. 

After determining the horizontal and
vertical magnitudes, the orientations of blur patterns are estimated jointly
considering the two orthogonal magnitudes. More specifically, MKSP, after a 1D (for $n = 1$) or 2D (for the rest) convolutional layer, fuses each pair of tensors (${\bf y}^{v,n}$, ${\bf y}^{h,n}$) into a tensor ${\bf y}^n \in R^{H \times W \times C}$ by
\begin{equation}
y^n_{i,j,c} = y^{v,n}_{i,\lfloor\frac{n\times j}{W}\rfloor,c} + y^{h,n}_{\lfloor\frac{n\times i}{H}\rfloor,j,c}.
\end{equation}

\begin{figure}[t]
    \centering
    \includegraphics[width=0.97\columnwidth]{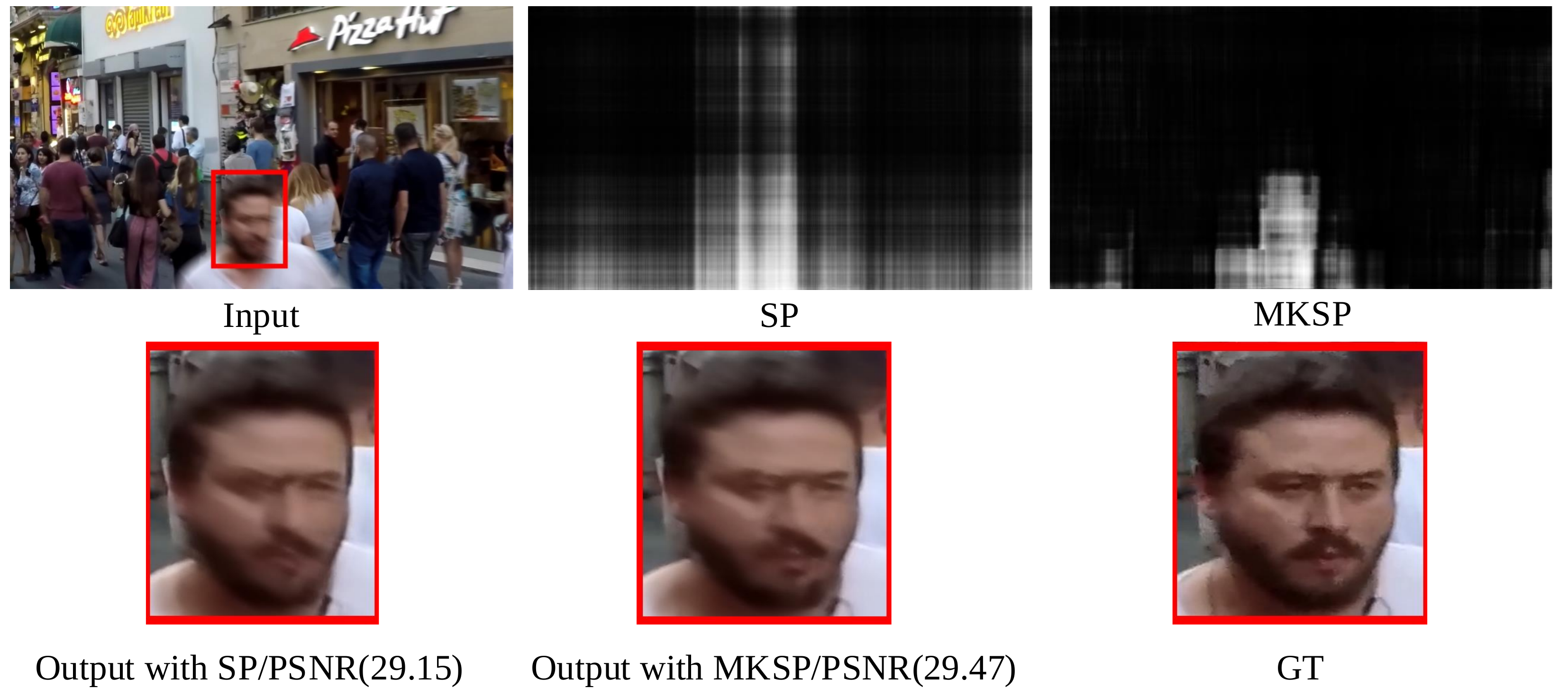}
    \caption{Visualization of the attention masks of SP and MKSP, and the corresponding effects on the final results on GoPro test set.}
    \label{fig:compared to sp}
\end{figure}

\noindent Similar to SP, we concatenate all the fused tensors to yield an attention mask as ${\bf M}_{mksp}= f_{out}({\bf y}^1\oplus{\bf y}^3 \oplus{\bf y}^5 \oplus{\bf y}^7)$, where $\oplus$ stands for the concatenation operation, and $f_{out}(\cdot)=\sigma_{sig}(Conv(\sigma_{ReLU}(Conv(\cdot))))$ represents a non-linear mapping function consisting of two $3\times 3$ convolutional layers. The first layer uses the ReLU activation function, and the second uses a sigmoid function. As shown in Fig.~\ref{fig:compared to sp}, the proposed MKSP can generate attention masks that better fit objects or local scenes than those by using SP with only $H\times 1$ and $1\times W$ kernels used, which yields rough band-shape masks.

\begin{figure}[t]
\centering
\includegraphics[width=1\columnwidth]{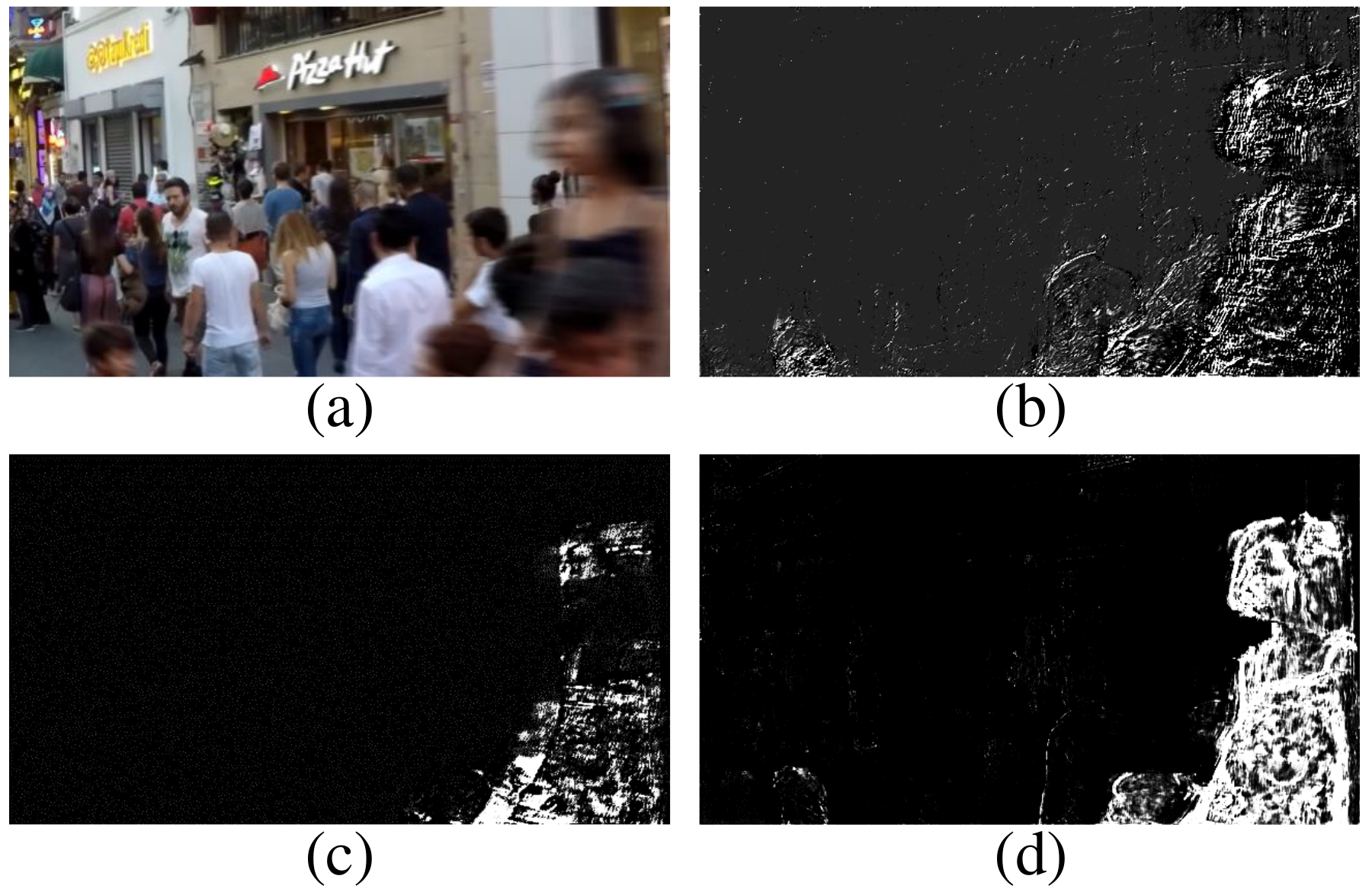}
\caption{(a) Input blurred image in GoPro testing set. (b)--(d) Comparisons among the attended feature maps by using different components of the proposed BA including (b) AR, (c) MKSP, and (d) MKSP + AR.}
\label{fig:baa mask}
\end{figure}

% yylin
\paragraph{Attention Refinement (AR)}
After obtaining the globally attended features by the element-wise multiplication of attention masks ${\bf M}_{mksp}$ and input tensor ${\bf x}$, we further refine these features locally via a simple attention mechanism using $f_{AR}(\cdot)$. The final output of our BA block through the MKSP and AR stages is computed as
\begin{equation}
 f_{AR}({\bf \Tilde{x}})\otimes{\bf \Tilde{x}},
\end{equation}
where $\otimes$ represents element-wise multiplication, and ${\bf \Tilde{x}}={\bf M}_{mksp} \otimes {\bf x}$ denotes the global features extracted using MKSP.
Figs.~\ref{fig:baa mask}(c) and (d) demonstrate that cascading MKSP with AR can refine the attended feature maps.

\begin{figure}[t]
\centering
\includegraphics[width=1.0\columnwidth]{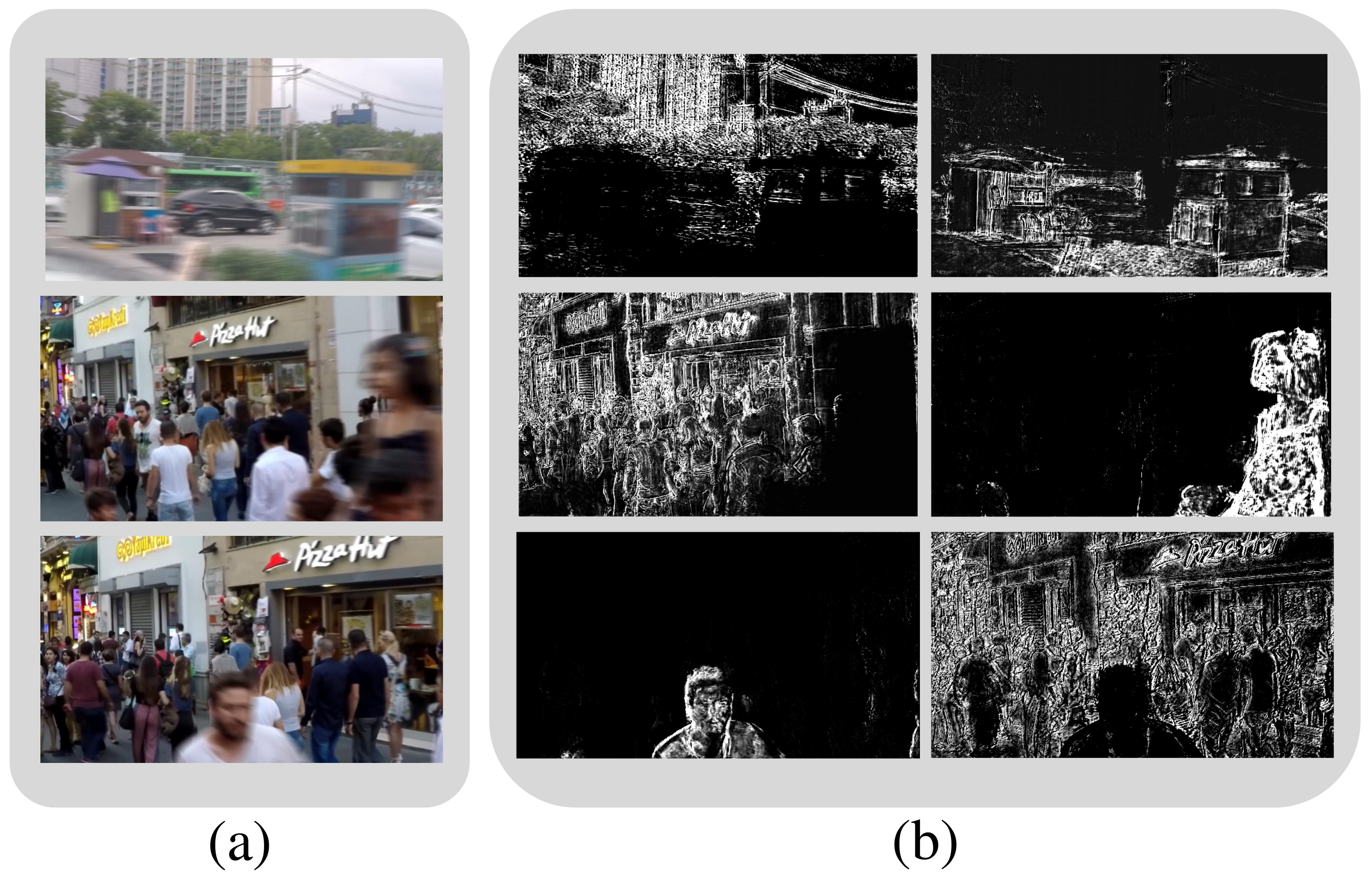}
\caption{Three disentanglement examples of blurred patterns of different degrees using our BA on GoPro test set. (a) Input blurred images and (b) attended feature maps on different regions.}
\label{fig:baa mask2}
\vspace{-0.2in}
\end{figure}

\begin{figure*}[t!]
\centering
\includegraphics[width=2\columnwidth]{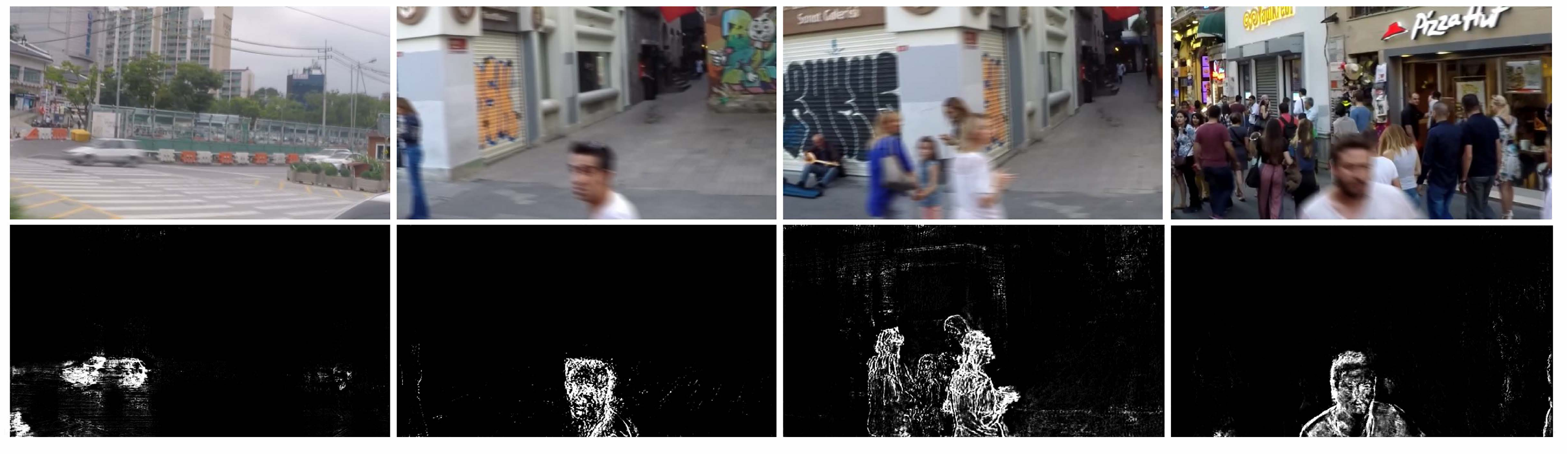}
\vspace{-0.1in}
\caption{Visualization of the blur-aware attended features on GoPro test set, where moving objects in the blurred images are highlighted while background is mostly excluded. These blur-aware masks are crucial for handling blurred images with diverse blur patterns.} 
\label{fig:attention3}
\end{figure*}

The proposed BA facilitates the attention mechanism applied to deblurring since it requires less memory, \ie $\mathcal{O}(HWC)$, where $C$ represents the channel dimension, than those adopted in~\cite{RADN_2020_ECCV, SAPN2020}. It disentangles blurred contents with different magnitudes and orientations. Fig.~\ref{fig:baa mask2} showcases three examples of blur content disentanglement using BA, where we witness that background scenes are differentiated from the foreground scenes because those objects closer to the camera move faster, thus more blurred. Fig.~\ref{fig:attention3} shows more examples of attention maps yielded by BA, which implicitly acts as a gate for propagating relevant blur contents. 

% yylin
\subsection{Cascaded Parallel Dilated Convolution (CPDC)}\label{subsec:CPDC}

\begin{figure}[t]
\centering
\includegraphics[width=1\columnwidth]{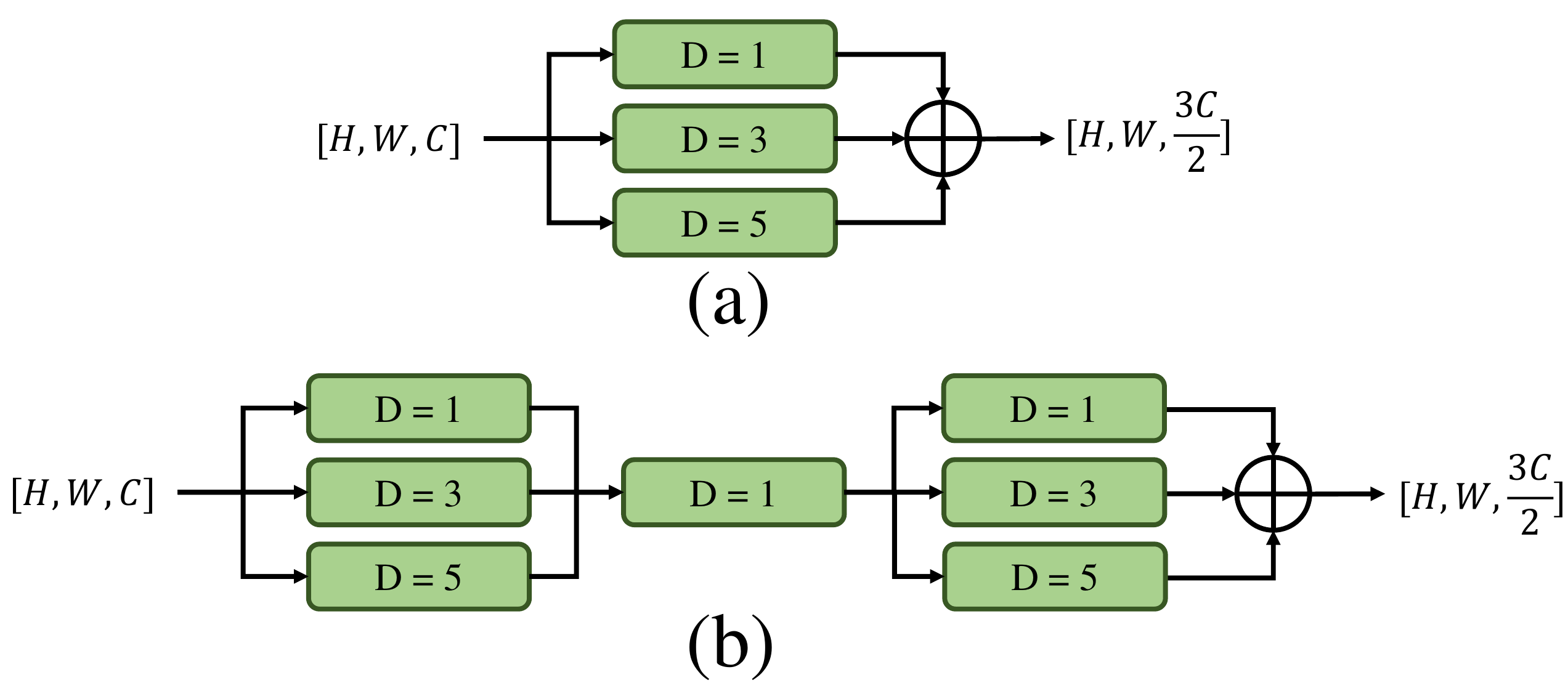}
\caption{Architectures of (a) parallel dilated convolution (PDC) and (b) cascaded parallel dilated convolution (CPDC).}
\label{fig:CPDC}
\end{figure}

Atrous convolution, also called {\em dilated convolution}, has been widely applied to computer-vision tasks~\cite{chen2017rethinking, Liu_2018_ECCV} for enlarging receptive fields and extracting features from objects with different scales without increasing the kernel size. 
Inspired by this, we design a {\em cascaded parallel dilated convolution} (CPDC) block with multiple dilation rates to capture multi-scale blurred objects. Instead of stacking dilated convolutional layers with different rates in parallel, which we call {\em parallel dilated convolution} (PDC), our CPDC block cascades two sets of PDC with a single convolutional layer working as a fusion bridge.
%channel attention bridge. 
It can distill patterns more beneficial to deblurring before passing through the second PDC. As an example, Fig.~\ref{fig:CPDC}(a) shows a PDC block consisting of three $3\times 3$ dilated convolutional layers with a dilation rate $D$ ($D=1, 3,$ and $5$),  each of which outputs features with half the number of input channels. After concatenation, the number of the output channels of the PDC block increases by $1.5$ times. As shown in Fig.~\ref{fig:CPDC}(b), our CPDC block consists of two PDC blocks bridged by a $3\times 3$ convolutional layer, which would be more effective in aggregating multi-scale content information for deblurring.

\subsection{Loss function}\label{subsec:loss}
In BANet, we utilize the Charbonnier loss as suggested in~\cite{Lai_2017_CVPR, Zamir_2021_CVPR}:
\begin{equation}
L_{char} = \sqrt{||\bf{R} - \bf{Y}||^2 + {\varepsilon}^2},
\end{equation}
where $\bf{R}$ and $\bf{Y}$ respectively denote the restored image and the ground-truth image, and $\varepsilon={10}^{-3}$ as in~\cite{Lai_2017_CVPR, Zamir_2021_CVPR}.
In addition, to enhance the restoration performance, we add an FFT loss to supervise the results in the frequency domain, as adopted in MIMO-UNet+~\cite{MIMO}:
\begin{equation}
L_{FFT} = ||\mathcal{F}(\bf{R})-\mathcal{F}(\bf{Y})||_1,
\end{equation}
where $\mathcal{F}$ represents the fast Fourier transform function. 
At last, we optimize BANet using the total loss $L$ as
\begin{equation}
 L = L_{char} + \lambda L_{FFT},
\end{equation}
where $\lambda$ is set to $0.01$ empirically.

% yylin
\section{Experiments}\label{4}
This section evaluates the proposed method. In the following, we first describe the experimental setup, then compare our method with the state-of-the-arts, and finally conduct ablation studies to analyze the effectiveness of individual components.\footnoteonlytext{The authors from the universities in Taiwan completed the experiments on the datasets.}
%charles
\vspace{-0.2in}

\begin{table}[t!]
\centering
\setlength{\tabcolsep}{1mm}
\caption{Evaluation results on GoPro test set. The best score in its column is highlighted in bold and the second best is underlined. Symbol $*$ indicates those methods without released code; thus we cite the results from the original papers or evaluate on the released deblurred images. All methods are trained on GoPro training set. Time and Params are measured in millisecond (ms) and million (M).}
%\vspace{-0.1in}
\begin{tabular}{l|c|c|c|c|c}\hline\hline
Model & PSNR $\uparrow$  & SSIM $\uparrow$ & Time $\downarrow$ & Params $\downarrow$ & GFLOPs $\downarrow$ \\\hline
MSCNN~\cite{Nah_2017_CVPR}  & 30.40 & 0.936 & 943 & 12 &  336    \\
SRN~\cite{tao2018srndeblur} & 30.25 & 0.934 & 650 & 7 & 167 \\
DSD~\cite{gao2019dynamic} & 30.96 & 0.942 & 1300 & \underline3 & 471 \\
DeblurGAN-v2~\cite{Kupyn_2019_ICCV} & 29.55 & 0.934 & 42 & 68 & \bf42\\
DMPHN~\cite{Zhang_2019_CVPR}  & 31.36 & 0.947 & 354 & 22 & 235\\
EDSD$^*$~\cite{Yuan_2020_CVPR} & 29.81 & 0.934 & \bf10 & \bf1 & -- \\
MTRNN~\cite{MT_2020_ECCV} & 31.13 & 0.944 & 53 & \underline3 & 164 \\
RADN$^*$~\cite{RADN_2020_ECCV} & 31.85 & 0.953 & 38 & -- & -- \\
SAPHN$^*$~\cite{SAPN2020} & 32.02 & 0.953 & 770 &  -- & --   \\
MIMO-UNet+~\cite{MIMO} & 32.45 & 0.957 & \underline{23} & 16 & \underline{154} \\
%MIMO (TTA)~\cite{MIMO} & 32.68 & \underline{0.959} & 30 & 16 & \underline{154} \\
MPRNet~\cite{Zamir_2021_CVPR} & \underline{32.66} & \underline{0.959} &  138 & 20  & 760    \\
\noalign{\hrule height 1.0pt}
BANet & 32.54 & 0.957 & \underline{23} & 18  & 264       \\
%BANet (TTA) & \underline{32.73} & \underline{0.959} & 35 & 18  & 264       \\
BANet+ & \bf33.03 & \bf0.961  & 25 & 40 & 588  \\\hline\hline
\end{tabular}
\label{Tab:Quant_eval}
\vspace{-0.1in}
\end{table}

\subsection{Experimental Setup}
We evaluate the BANet on three image deblurring benchmark datasets: 1) GoPro~\cite{Nah_2017_CVPR} that consists of $3,214$ pairs of blurred and sharp images of resolution $720\times 1280$, where $2,103$ pairs are used for training, and the rest for testing, 2) HIDE~\cite{HAdeblur} that contains $2,025$ pairs of HD images, all for testing, and RealBlur~\cite{rim_2020_ECCV} that consists of $3,758$ pairs for training and $980$ pairs for testing. 
The RealBlur dataset is further split into two datasets: RealBlur-R collected from raw images and RealBlur-J from JPEG images.
We train our model using Adam optimizer with parameters $\beta_{1}=0.9$ and $\beta_{2}=0.999$.
We set the initial learning rate to $10^{-4}$, which then decays to $10^{-7}$ based on the cosine annealing strategy.
Following~\cite{MT_2020_ECCV, Yuan_2020_CVPR}, we utilize random cropping, flipping, and rotation for data augmentation. 
Lastly, we implement our model with PyTorch library on a computer equipped with Intel Xeon Silver 4210 CPU and NVIDIA 2080ti GPU.

\begin{figure*}[t!]
\centering
\includegraphics[width=2\columnwidth]{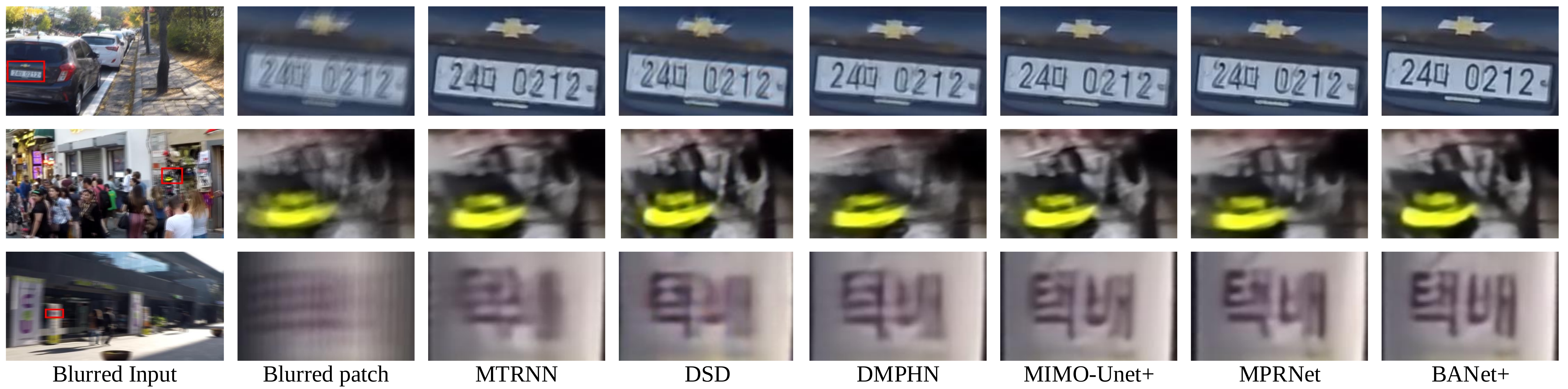}
\vspace{-0.1in}
\caption{Qualitative comparisons on GoPro~\cite{Nah_2017_CVPR} test set. The deblurred results listed from left to right are from MTRNN~\cite{MT_2020_ECCV}, DSD~\cite{gao2019dynamic},  DMPHN~\cite{Zhang_2019_CVPR}, MIMO-UNet+\cite{MIMO}, MPRNet~\cite{Zamir_2021_CVPR}, and Ours.}
\label{fig:Visualization_Result_GoPro}
%\vspace{-0.1in}
\end{figure*}

\begin{table}[t!]
\centering
\setlength{\tabcolsep}{4.5mm}
\caption{Evaluation results on HIDE dataset. The best score in its column is highlighted in bold and the second best is underlined. Symbol $*$ indicates those methods without released code; thus we cite the results from the original papers or evaluate on the released deblurred images. All methods are trained on GoPro training set. Time is measured in millisecond (ms).}
%\vspace{-0.1in}
\begin{tabular}{l|c|c|c}\hline\hline
{Model} & {PSNR $\uparrow$}  & {SSIM $\uparrow$} & Time $\downarrow$ \\\hline
DeblurGAN-v2~\cite{Kupyn_2019_ICCV} & 27.40 & 0.882 & 42 \\
SRN~\cite{tao2018srndeblur} & 28.36 & 0.904 & 424 \\
HAdeblur$^*$~\cite{HAdeblur} & 28.87 & 0.903 & -- \\
DSD~\cite{gao2019dynamic} & 29.10 & 0.913 & 1200 \\
DMPHN~\cite{Zhang_2019_CVPR} & 29.10 & 0.918 & 341 \\
MTRNN~\cite{MT_2020_ECCV} & 29.15 & 0.918 & 53 \\
SAPHN$^*$~\cite{SAPN2020}  & 29.98 & 0.930 & -- \\
MIMO-UNet+~\cite{MIMO} & 30.00 & 0.930 & 28     \\
%MIMO (TTA)~\cite{MIMO} & 30.45 & 0.935 &  & 16      \\
MPRNet~\cite{Zamir_2021_CVPR} & \bf30.93 & \bf0.939 & 138 \\
\noalign{\hrule height 1.0pt}
BANet   & 30.16 & 0.930 & \bf23 \\
%BANet (TTA)  & 30.42 & 0.933 &  & 18       \\
BANet+ & \underline{30.58} & \underline{0.935} & \underline{25} \\\hline\hline
\end{tabular}
\label{Tab:Quant_eval_HIDE}
%\vspace{-0.2in}
\end{table}

\begin{figure*}[t!]
\centering
\includegraphics[width=2\columnwidth]{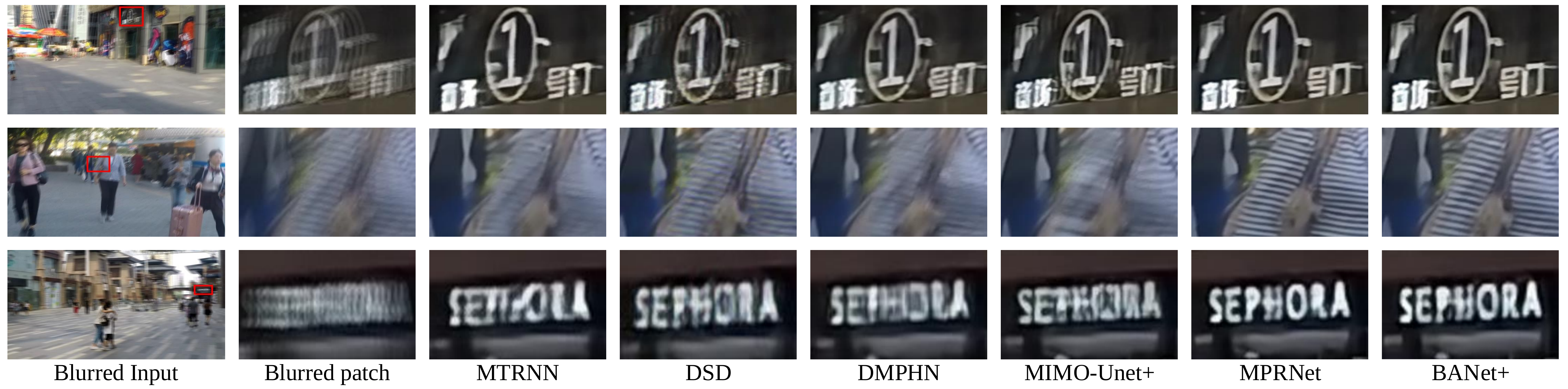}
\vspace{-0.1in}
\caption{Qualitative comparisons on HIDE~\cite{HAdeblur} dataset. The deblurred results listed from left to right are from MTRNN~\cite{MT_2020_ECCV}, DSD~\cite{gao2019dynamic},  DMPHN~\cite{Zhang_2019_CVPR}, MIMO-UNet+\cite{MIMO}, MPRNet~\cite{Zamir_2021_CVPR}, and Ours.}
\label{fig:Visualization Result_HIDE}
%\vspace{-0.2in}
\end{figure*}

\vspace{-0.1in}
\subsection{Experimental Results}
\noindent\textbf{Quantitative Analysis}:
%{\bf Quantitative Evaluation:}
We compare our method with 11 latest approaches, including MSCNN~\cite{Nah_2017_CVPR}, SRN~\cite{tao2018srndeblur}, DSD~\cite{gao2019dynamic}, DeblurGAN-v2~\cite{Kupyn_2019_ICCV}, DMPHN~\cite{Zhang_2019_CVPR},
EDSD~\cite{Yuan_2020_CVPR}, 
MTRNN~\cite{MT_2020_ECCV}, RADN~\cite{RADN_2020_ECCV},
SAPHN~\cite{SAPN2020},
MIMO-UNet+~\cite{MIMO}, and MPRNet~\cite{Zamir_2021_CVPR}, which also handle dynamic deblurring on the GoPro~\cite{Nah_2017_CVPR} test set. 
For HIDE~\cite{HAdeblur}, we choose nine recent deblurring methods, including DeblurGAN-v2~\cite{Kupyn_2019_ICCV}, SRN~\cite{tao2018srndeblur}, HAdeblur~\cite{HAdeblur}, DSD~\cite{gao2019dynamic}, DMPHN~\cite{Zhang_2019_CVPR}, MTRNN~\cite{MT_2020_ECCV}, SAPHN~\cite{SAPN2020}, MIMO-UNet+~\cite{MIMO}, and MPRNet~\cite{Zamir_2021_CVPR}, according to their availability in released pre-trained weights.
For RealBlur~\cite{rim_2020_ECCV}, we choose four methods that trained on the RealBlur training set, including DeblurGAN-v2~\cite{Kupyn_2019_ICCV}, SRN~\cite{tao2018srndeblur}, MPRNet~\cite{Zamir_2021_CVPR}, and MIMO-UNet+~\cite{MIMO}.

To better compare with recent approaches, we devise two versions of our model, BANet and BANet+. The only difference between them is the number of channels used in a BAM, and BANet with 128 channels involves 18 million parameters while BANet+ has 40 million parameters with 192 channels.
Table~\ref{Tab:Quant_eval} lists the objective scores (PSNR and SSIM), runtime, parameters, and GFLOPs on the GoPro test set for all the compared methods.
We observe that the self-recurrent models, MSCNN~\cite{Nah_2017_CVPR}, SRN~\cite{tao2018srndeblur}, DSD~\cite{gao2019dynamic}, MTRNN~\cite{MT_2020_ECCV}, SAPHN~\cite{SAPN2020}, and MPRNet~\cite{Zamir_2021_CVPR}, consume longer runtime than the non-recurrent ones,~\ie,~DeblurGAN-v2~\cite{Kupyn_2019_ICCV}, RADN~\cite{RADN_2020_ECCV}, MIMO-UNet+~\cite{MIMO}, and ours.
%
%We record the average runtime of all the models using a single GPU. 
%
As reported in Table~\ref{Tab:Quant_eval}, BANet runs faster with fewer parameters and GFLOPs as well as achieves better performance than recurrent-based methods, MSCNN~\cite{Nah_2017_CVPR}, SRN~\cite{tao2018srndeblur}, DSD~\cite{gao2019dynamic}, DMPHN~\cite{Zhang_2019_CVPR}, MTRNN~\cite{MT_2020_ECCV}, and SAPHN~\cite{SAPN2020} and non-recurrent methods, such as DeblurGAN-v2~\cite{Kupyn_2019_ICCV} and RADN~\cite{RADN_2020_ECCV} on the GoPro test set. BANet also performs favorably against an efficient multi-scale model, MIMO-UNet+~\cite{MIMO}, with the same runtime and a comparable model size. 
BANet+ outperforms the best competitor, MPRNet~\cite{Zamir_2021_CVPR}, by 0.37 dB in PSNR with faster runtime ($-113$ms) and lower GFLOPs ($-172$).
Table~\ref{Tab:Quant_eval_HIDE} shows the quantitative results on HIDE~\cite{HAdeblur}. As can be seen, BANet outperforms all the compared methods except for MPRNet~\cite{Zamir_2021_CVPR} with a faster inference time. BANet+ only works comparably to MPRNet~\cite{Zamir_2021_CVPR} since MPRNet seems to perform favorably on HIDE~\cite{HAdeblur} particularly, but our model runs much faster.
Table~\ref{Tab:Quant_eval_RealBlur} lists the quantitative comparisons on the RealBlur test set, demonstrating that both BANet and BANet+ outperform the compared methods on the RealBlur-J and RealBlur-R test sets.

\begin{figure*}[t!]
%\vspace{1.5in}
\centering
\includegraphics[width=2\columnwidth]{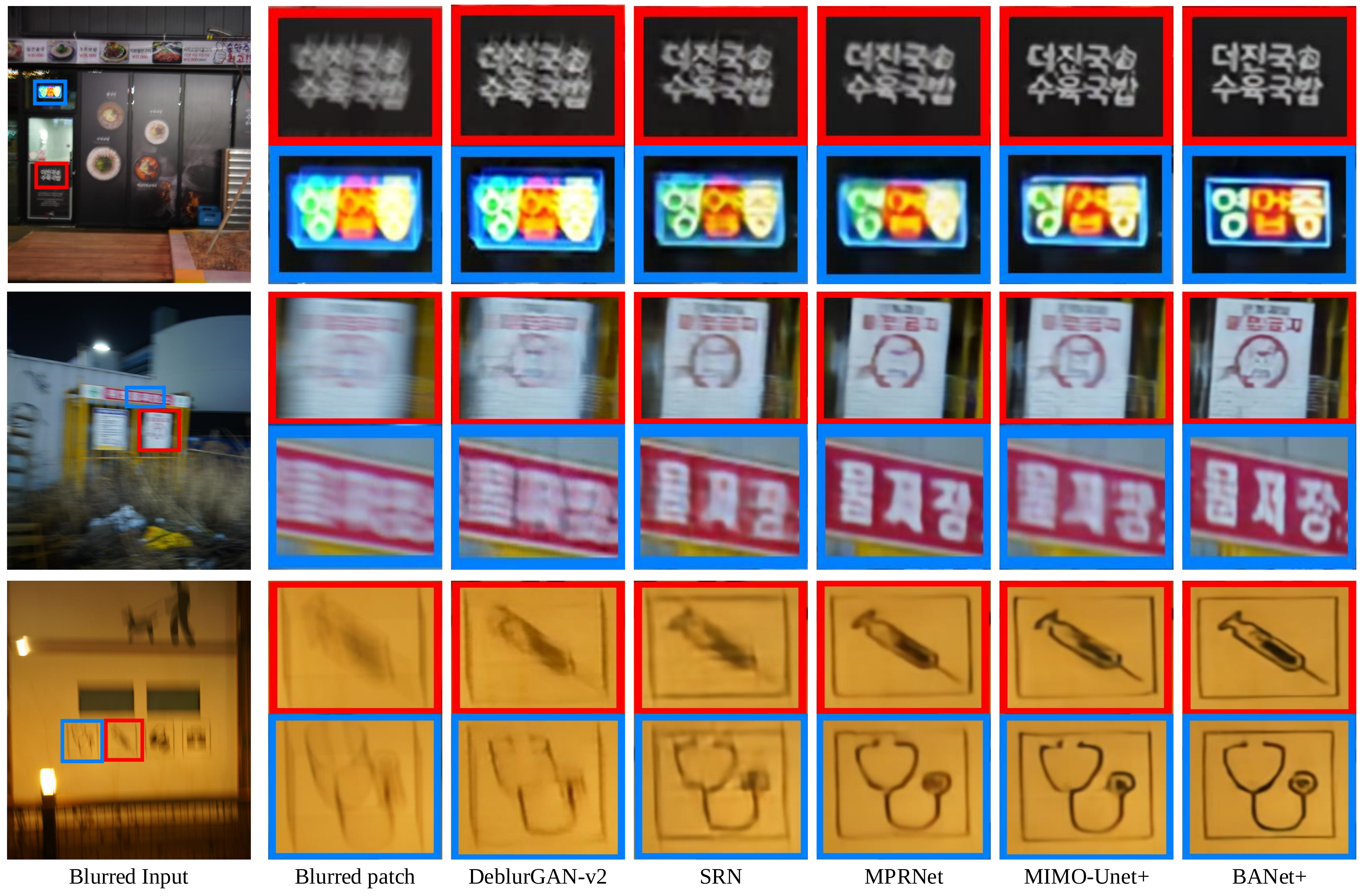}
\vspace{-0.1in}
\caption{Examples of deblurred results obtained using DeblurGAN-v2~\cite{Kupyn_2019_ICCV}, SRN~\cite{tao2018srndeblur},  MPRNet~\cite{Zamir_2021_CVPR}, MIMO-UNet+~\cite{MIMO}, and Ours on RealBlur~\cite{rim_2020_ECCV} test set.}
\label{fig:RealBlur_A}
\vspace{-0.2in}
\end{figure*}

\begin{table}[t!]
\centering
\setlength{\tabcolsep}{1.5mm}
\caption{Evaluation results on RealBlur test set. All methods are trained on RealBlur training set. Time is measured in millisecond (ms).}
\vspace{-0.1in}
\begin{tabular}{l|cc|cc|c}
\hline\hline
& \multicolumn{2}{c|}{RealBlur-J} & \multicolumn{2}{c|}{RealBlur-R} &\\ %\multicolumn{1}{c}{RealBlur} \\
Model & PSNR $\uparrow$ & SSIM $\uparrow$  & PSNR $\uparrow$ & SSIM $\uparrow$ & Time \\ \hline
DeblurGAN-v2~\cite{Kupyn_2019_ICCV} & 29.69    & 0.870  & 36.44  & 0.935 & 44 \\
SRN~\cite{tao2018srndeblur} & 31.38  & 0.901 & 38.65  & 0.965 & 420 \\
MPRNet~\cite{Zamir_2021_CVPR} & 31.76 & 0.922 & 39.31 & \bf0.972 & 81  \\
MIMO-UNet+~\cite{MIMO} & 31.92 & 0.919 & -- & -- & \underline{23} \\
\noalign{\hrule height 1.0pt}
BANet  & \underline{32.00} & \underline{0.923}  & \underline{39.55}  & \underline{0.971}  &  \bf{22} \\
BANet+  & \bf{32.42}  & \bf0.929 & \bf39.90  & \bf{0.972}  &  24  \\
\hline\hline
\end{tabular}
\label{Tab:Quant_eval_RealBlur}
%\vspace{-0.25in}
\end{table}

\begin{figure*}[t!]
\centering
\includegraphics[width=2\columnwidth]{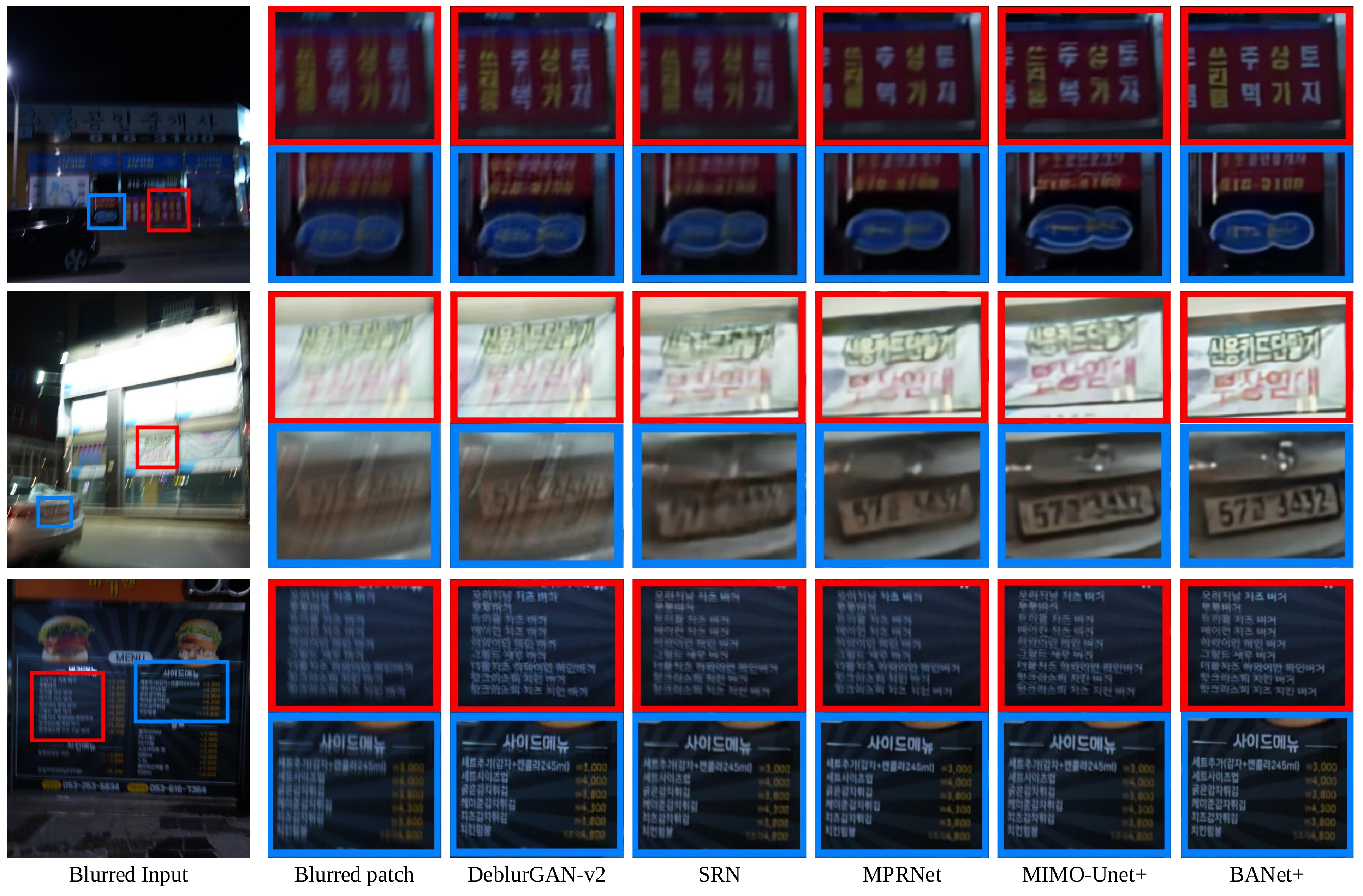}
\vspace{-0.1in}
\caption{Examples of deblurred results obtained using DeblurGAN-v2~\cite{Kupyn_2019_ICCV}, SRN~\cite{tao2018srndeblur},  MPRNet~\cite{Zamir_2021_CVPR}, MIMO-UNet+~\cite{MIMO}, and Ours on RealBlur~\cite{rim_2020_ECCV} test set.}
\label{fig:RealBlur_B}
\vspace{-0.1in}
\end{figure*}

\noindent\textbf{Qualitative Analysis}:
Fig.~\ref{fig:Visualization_Result_GoPro} and Fig.~\ref{fig:Visualization Result_HIDE} show qualitative comparisons on the GoPro test set and HIDE dataset with previous state-of-the-arts MTRNN~\cite{MT_2020_ECCV}, DSD~\cite{gao2019dynamic}, DMPHN~\cite{Zhang_2019_CVPR}, MIMO-UNet+~\cite{MIMO}, and MPRNet~\cite{Zamir_2021_CVPR}.
As observed in Fig.~\ref{fig:Visualization_Result_GoPro}, MTRNN~\cite{MT_2020_ECCV}, DSD~\cite{gao2019dynamic}, DMPHN~\cite{Zhang_2019_CVPR}, MIMO-UNet+~\cite{MIMO}, and MPRNet~\cite{Zamir_2021_CVPR} do not well recover regions with texts or severe blurs whereas BANet can restore those regions better.
In Fig.~\ref{fig:Visualization Result_HIDE}, MTRNN~\cite{MT_2020_ECCV}, DSD~\cite{gao2019dynamic}, DMPHN~\cite{Zhang_2019_CVPR}, and MIMO-UNet+~\cite{MIMO} do not deblur the striped t-shirt and texts well, while BANet recovers those parts better. 
Fig.~\ref{fig:RealBlur_A} and Fig.~\ref{fig:RealBlur_B} demonstrate some deblurred results using DeblurGAN-v2~\cite{Kupyn_2019_ICCV}, SRN~\cite{tao2018srndeblur}, MPRNet~\cite{Zamir_2021_CVPR}, MIMO-UNet+~\cite{MIMO}, and ours, on the RealBlur~\cite{rim_2020_ECCV} test set. As can be seen, although all these models can remove blurs, BANet performs favorably on delicate image details.
%\vspace{-0.1in}

\noindent\textbf{User Study}:
We further conduct a user study to evaluate the subjective quality of deblurred results on real blurred images chosen from the RealBlur-J test set. We compare our method (BANet+) against four methods, including MIMO-UNet+~\cite{MIMO}, MPRNet~\cite{Zamir_2021_CVPR}, SRN~\cite{tao2018srndeblur}, and DeblurGAN-v2~\cite{Kupyn_2019_ICCV}. Note that all the methods are trained on the RealBlur-J training set.

In the study, 34 subjects aged from 21 to 40 years participated in the study without any prior knowledge of the experiment. Their vision is either normal or corrected to be normal. We picked $16$ blurred images with varying scenes for the experiment and obtained the deblurred results using all the compared approaches.
Since each method is compared against BANet with all the chosen blurred images in the experiment, we have $16\times4=64$ image pairs in total. Each subject is shown all the image pairs, one at a time, and asked which one he/she prefers in terms of visual quality. Each image pair is displayed randomly and placed side by side. Subjects are asked to check images carefully before choosing without a time limit.

Table~\ref{tab:user study} shows the subjective evaluation results, where the values represent the percentage that the deblurring results with our method are preferred to the counterparts with the other compared methods for all the votes collected. It indicates that our method obtains over 95\% preference votes compared to all the compared methods, which again demonstrates that our approach achieves better subjective visual quality.

\begin{table}[t]
\centering
\setlength{\tabcolsep}{1mm}
\caption{Results of User Study. The values represent the percentage that our method was chosen over the other compared methods.}
\begin{tabular}{c|cccc}\hline
 & MIMO-UNet+~\cite{MIMO} & MPRNet~\cite{Zamir_2021_CVPR}  & SRN~\cite{tao2018srndeblur} & DeblurGAN-v2~\cite{Kupyn_2019_ICCV} \\\hline
Ours & 95.6\% & 95.6\%  & 98.5\% &  99.6\% \\\hline 
\end{tabular}
\label{tab:user study}
\vspace{-0.1in}
\end{table}

\subsection{Ablation Study}
In the ablation studies, we do all the experiments on the BANet (18M) version.

\noindent{\bf BAM with Different Components:}
Table~\ref{tab:BAM_ablation} shows an ablation on different component combinations in our Blur-Aware Module (BAM) tested on the GoPro test set. As can be seen, adding a simple attention refinement (AR) mechanism to PDC (Net1~vs.~Net2) can improve PSNR by 0.41 dB, which shows the effectiveness of spatial attention for deblurring.
Using MKSP in PDC (Net1~vs.~Net4) improves PSNR by 0.72 dB, which has much more performance gain compared to using strip pooling (SP)~\cite{hou2020strip} in Net3 or AR in Net2.   
Substituting PDC in Net5 with CPDC (Net6), our proposed version of BAM leads to a further performance gain.
Thanks to its mechanism for locating blur regions based on both global attention and local convolutions, our BAM attains the best performance while achieving fast inference time.    

\begin{table}[t!]
\centering
\setlength{\tabcolsep}{3mm}
\caption{Ablation study on GoPro test set using different component combinations in BAM}
\begin{tabular}{c|cccccc}\hline
Model & PDC & AR & SP &  MKSP & CPDC & PSNR \\\hline
Net1  & $\surd$ &   &  &  & & 31.39   \\
Net2  & $\surd$ & $\surd$ & & & & 31.80  \\
Net3  & $\surd$ & & $\surd$ & & & 31.81  \\
Net4  & $\surd$ & &  & $\surd$ & & 32.11  \\
Net5  & $\surd$ & $\surd$ &  & $\surd$ & & 32.24  \\
Net6  & & $\surd$ &  & $\surd$ & $\surd$ & 32.54  \\
\hline
\end{tabular}
\label{tab:BAM_ablation}
%\vspace{-0.2in}
\end{table}

\noindent{\bf Numbers of Stacked BAMs:}
%charles
Using more layers to enlarge the receptive field may improve performance for computer vision or image processing tasks. 
Nevertheless, stacking more layers for deblurring does not guarantee better performance~\cite{SAPN2020} and might consume extra inference time. 
However, using our residual learning-based BAM design, we can stack multiple layers to expand the effective receptive field for better deblurring. 
In Table~\ref{tab:BAM_ablation2}, we show performance comparisons with various numbers of BAMs stacked in our model on the GoPro test set. 
We list four versions: stack-4, stack-8, stack-10, and stack-12, corresponding to 4, 8, 10, and 12 BAMs stacked in BANet. 
Although the quantitative performance improves with the number of  BAMs, the improvement became saturated after 12. 
Therefore, we choose 10 for its excellent balance between efficiency and visual quality.

\begin{table}[t]
\centering
\setlength{\tabcolsep}{4mm}
\caption{Performance comparisons of the stacking number of BAMs in BANet on GoPro test set.}
\begin{tabular}{c|cccc}\hline
BANet & stack-4 & stack-8  & stack-10 & stack-12 \\\hline
PSNR & 31.36 & 32.37  & 32.54 &  \bf32.55 \\\hline 
\end{tabular}
\label{tab:BAM_ablation2}
%\vspace{-0.2in}
\end{table}

\begin{table}[t]
\centering
\setlength{\tabcolsep}{3mm}
\caption{Performance comparisons of strip pooling (SP) and MKSP on GoPro test set with PDC.}
\begin{tabular}{c|cccc}\hline
BANet & SP & MKSP$_{135}$ & MKSP$_{1357}$ & MKSP$_{13579}$ \\\hline
PSNR & 31.81 & 32.03  & \bf32.11 &  32.04 \\\hline 
\end{tabular}
\label{tab:MKSP_ablation}
%\vspace{-0.2in}
\end{table}

\noindent{\bf Effectiveness of MKSP and CPDC:}
In Table~\ref{tab:MKSP_ablation}, we investigate the effects of kernel combination of MKSP on the GoPro test set. 
MKSP with five kernel sizes of $1$, $3$, $5$, $7$, and $9$ performs a little worse than
the first four sizes ($1$, $3$, $5$, and $7$), indicating that adding the kernel size of $9$ would not catch blur features more accurately, thus not helping with
the performance.
In Table~\ref{tab:compared CPDC with pdc}, we verify that CPDC, which uses a single convolution as a fusion bridge, outperforms PDC. 
For a fair comparison, we also compare CPDC against a PDC variant that stacks two PDCs in a series, called PDC$^2$, with a similar parameter size, and CPDC still performs better. 

\begin{table}[t]
\centering
\setlength{\tabcolsep}{6mm}
\caption{Ablation study of CPDC (w/o BA) compared to PDC (w/o BA) on GoPro test set.}
\begin{tabular}{cccc}\hline
 & PDC$_{135}$ & PDC$^2_{135}$ & CDPC \\\hline
PSNR & 31.39 & 31.78  & \bf32.13  \\ 
Parms (M) & \bf6 & 10  & 10  \\\hline
\end{tabular}
\label{tab:compared CPDC with pdc}
\end{table}

\iffalse
\begin{table}[t!]
\centering
\setlength{\tabcolsep}{3mm}
\caption{}
\begin{tabular}{cccc}\hline
& CPDC$_{13}$ & CDPC$_{135}$  & CPDC$_{1357}$ \\\hline
PSNR & 32.03 & 32.13  &  32.34   \\ 
Parms (M) & 6 & 10  & 14  \\\hline
\end{tabular}
\label{tab:CPDC with different dilated rate}
%\vspace{-0.2in}
\end{table}
\fi

\subsection{Blur-aware Attention vs. Self-Attention}
\begin{figure}[t!]
\centering
\includegraphics[width=1\columnwidth]{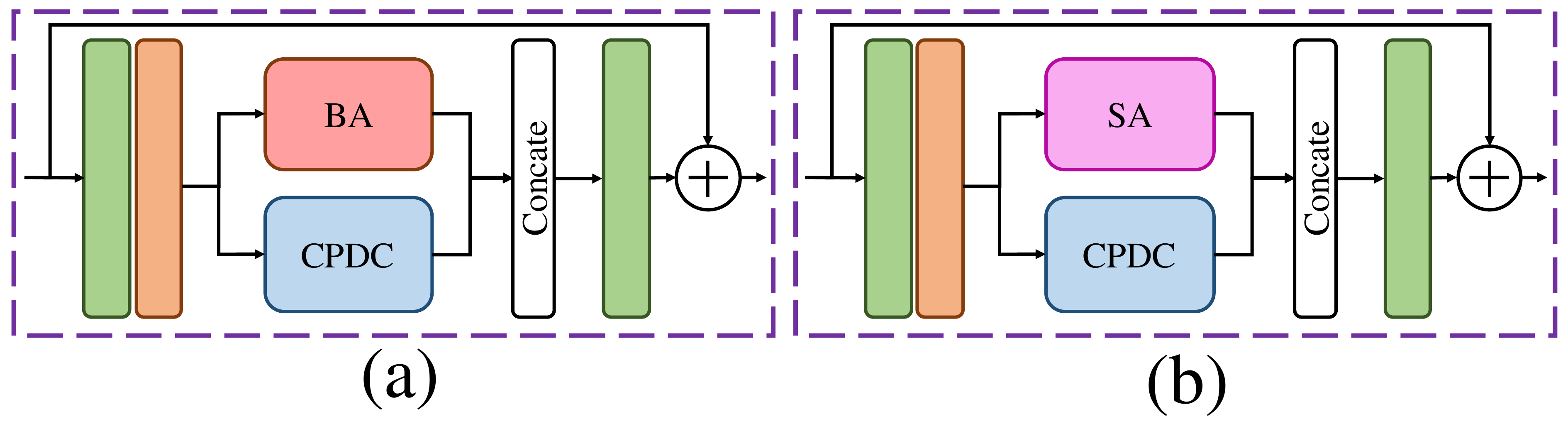}
%\vspace{-0.3in}
\caption{Architecture comparisons between (a) our original BAM and (b) BA replaced by SA~\cite{SAGAN_2019_PMLR} in BAM.}
\label{fig:compared with sa}
\end{figure}

RADN~\cite{RADN_2020_ECCV} utilizes a similar self-attention (SA) mechanism proposed in~\cite{SAGAN_2019_PMLR} for deblurring. 
It helps connect regions with similar blurs to facilitate global access to relevant features across the entire input feature maps. 
However, its high memory usage makes applying it to high-resolution images  infeasible. 
Thus, SA is usually employed in network layers on a smaller scale like in RADN~\cite{RADN_2020_ECCV}, where important blur information would be lost due to down-sampling. 
In contrast, our proposed region-based attention is more suitable for correlating regions with similar blur characteristics. Moreover, it can process high-resolution images thanks to its low memory consumption. 
To further demonstrate our BA’s efficacy, we compare the SA~\cite{SAGAN_2019_PMLR} with BA using our BANet (stack-4) as a backbone network, as shown in Fig.~\ref{fig:compared with sa}(b). 
Due to the high memory demand for SA ($\mathcal{O}(H^2W^2)$) to process $720\times 1280$ images, we adopt our stack-4 model for training. 
When testing the networks, we separate the input image into eight sub-images for both SA and BA to deblur, each equipped with a single 2080ti GPU. Since our BA requiring lower memory usage ($\mathcal{O}(CHW)$, where $C<<H \times W$) can process the image with the full resolution, we also show its result.
In Table~\ref{tab:SA_vs_BA}, SA$^{*}$ and BA$^{*}$ represent deblurring an image with its eight sub-images separately, whereas BA for processing the entire image at once. As can be observed, the proposed BA$^{*}$ works much more efficiently than SA$^{*}$ with a comparable result. When deblurring the entire image at once, BA undoubtedly performs the best.
%We provide the deblurring results using BA with or without splitting the input image into eight sub-images for a fair comparison. Table~\ref{tab:SA_vs_BA} shows the results under the scenario of dividing the input into eight sub-images, demonstrating BA still outperforms SA for deblurring based on our BANet. Also, dividing the input image for deblurring using a single GPU increases the run-time, and adopting BA runs significantly faster. Last but not least, since our BA has lower memory usage, we can process the original input image without division to attain better visual quality.  

\begin{table}[t!]
\centering
\setlength{\tabcolsep}{7mm}
%\vspace{-0.3in}
\caption{Performance comparison between BA and SA~\cite{SAGAN_2019_PMLR} using BANet (stack-4) on GoPro test set. $*$ represents deblurring on eight sub-images instead of an entire image.}
\begin{tabular}{cccc}\\\hline
    & SA$^{*}$ & BA$^{*}$ & BA  \\\hline
    PSNR &  31.11  & 31.09 & \bf31.36    \\
    Time (ms)  & 770 & 16 & \bf12 \\\hline   
\end{tabular}
\label{tab:SA_vs_BA}
\end{table}

\section{Conclusion}   
This paper proposes a novel blur-aware attention network (BANet) for single image deblurring. BANet consists of stacked blur-aware modules (BAMs) to disentangle region-wise blur contents of different magnitudes and orientations and aggregate multi-scale content features for more accurate and efficient dynamic scene deblurring. We have investigated and examined our design through demonstrations of attention masks and attended feature maps, as well as extensive ablation studies and performance comparisons. Our extensive experiments demonstrate that the proposed BANet achieves real-time deblurring and performs favorably against state-of-the-art deblurring methods on the GoPro and RealBlur benchmark datasets.
%\vspace{1.5in}

% \clearpage
% \newpage
{\small
\bibliographystyle{IEEEtran}
\bibliography{egbib}
}
\begin{IEEEbiography}
[{\includegraphics[width=1in,height=1.25in,clip,keepaspectratio]{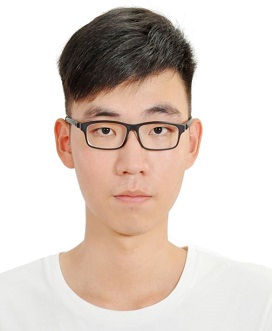}}]
{Fu-Jen~Tsai} received the B.S. degree in electrical engineering from National Taiwan Ocean University, Keelung, Taiwan, and the M.S. degree in communications engineering from National Tsing-Hua University, Hsinchu, Taiwan, in 2019 and 2021, respectively. Since 2021, he has been pursuing the Ph.D. degree in the department of electrical engineering, National Tsing-Hua University, Hsinchu, Taiwan. His research interests include image
/video restoration and computer vision. 

\end{IEEEbiography}

%\newpage

\vspace{-0.4in}
\begin{IEEEbiography}
	[{\includegraphics[width=1in,height=1.25in,clip,keepaspectratio] {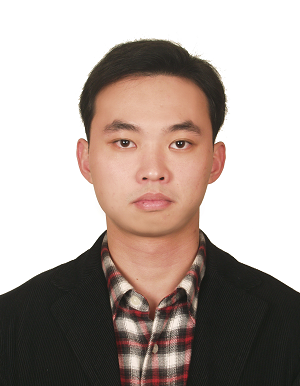}}]
	{Yan-Tsung Peng} (Member, IEEE) received the Ph.D. in electrical and computer engineering from University of California, San Diego, in 2017. He joined National Chengchi University in Feb. 2019, where he is currently an Assistant Professor in the computer science department. Before that, he was a Senior Engineer with Qualcomm Technologies, Inc., San Diego. His research interests include image processing, video compression, and machine-learning applications.

\end{IEEEbiography}

\vspace{-0.4in}
\begin{IEEEbiography}
	[{\includegraphics[width=1in,height=1.25in,clip,keepaspectratio] {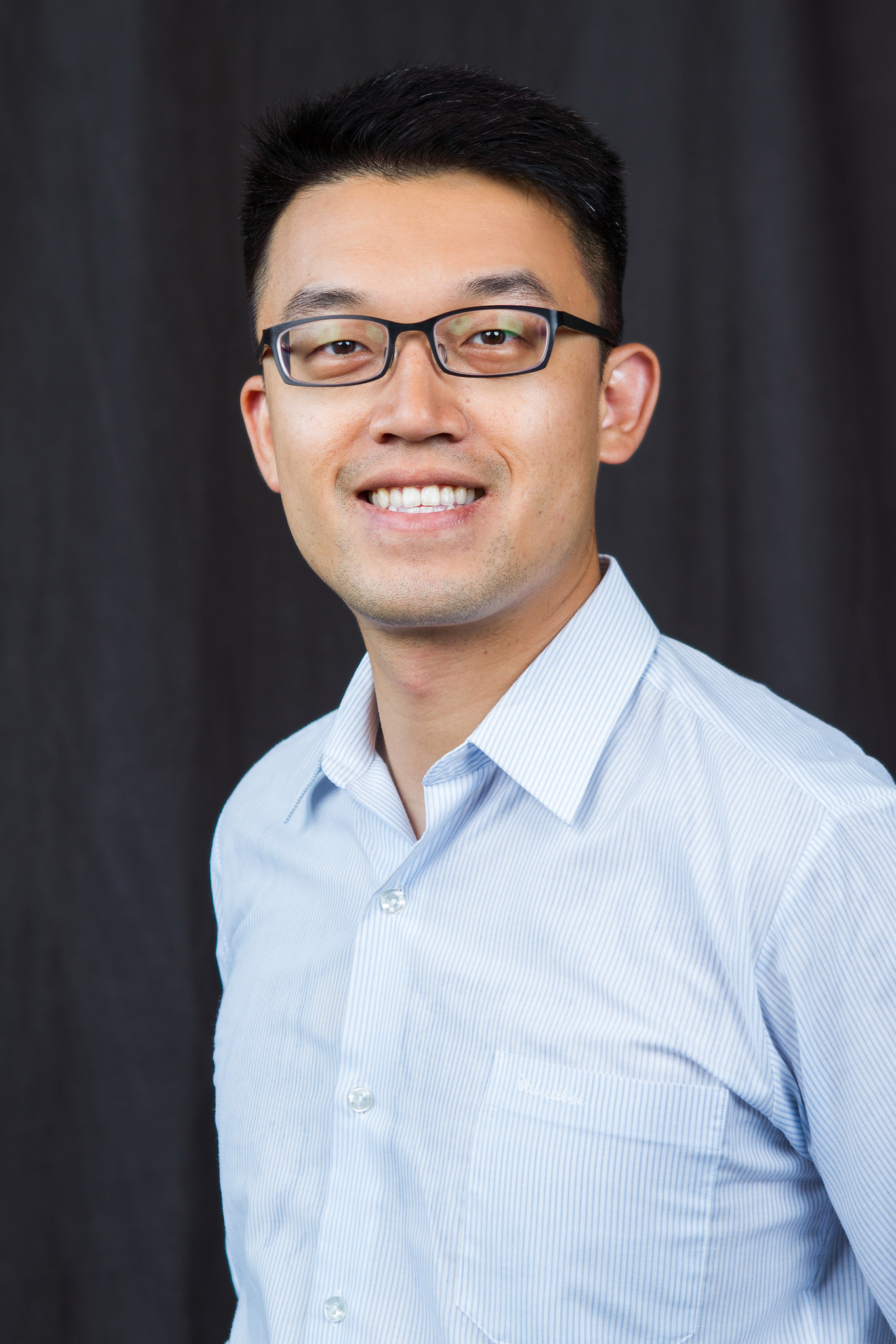}}]
	{Chung-Chi ``Charles'' Tsai} received his B.S. degree from National Tsing-Hua University, Hsinchu, Taiwan, and the M.S. degree from University of California at Santa Barbara, Santa Barbara, CA, USA, and the Ph.D. degree from Texas A\&M University, College Station, TX, USA, in 2009, 2012 and 2018, respectively and all in Electrical Engineering. He received full scholarship from Ministry of Education, Taiwan, to attend a one-year exchange program, at the University of New Mexico, Albuquerque, NM, USA, in 2007 and also participated in the summer internship with MediaTek in 2013/2015/2016. He is currently a staff engineer for camera machine learning team at Qualcomm Technologies, Inc., USA. His research interests include image processing and computer vision.

\end{IEEEbiography}

\vspace{-0.4in}
\begin{IEEEbiography}[{\includegraphics[width=1in,height=1.25in,clip,keepaspectratio]{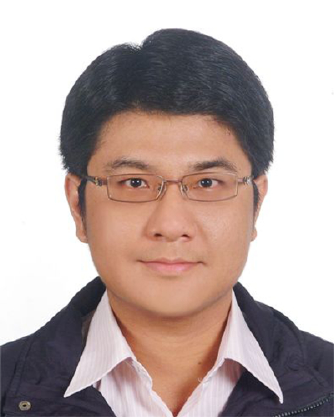}}]{Yen-Yu Lin} (Senior Member, IEEE) 
received the B.B.A. degree in Information Management, and the M.S. and Ph.D. degrees in Computer Science and Information Engineering from National Taiwan University, Taipei, Taiwan, in 2001, 2003, and 2010, respectively. He is currently a Professor with the Department of Computer Science, National Yang Ming Chiao Tung University, Hsinchu, Taiwan. His research interests include computer vision, machine learning, and artificial intelligence.
\end{IEEEbiography}

\vspace{-0.4in}
\begin{IEEEbiography}
	[{\includegraphics[width=1in,height=1.25in,clip,keepaspectratio] {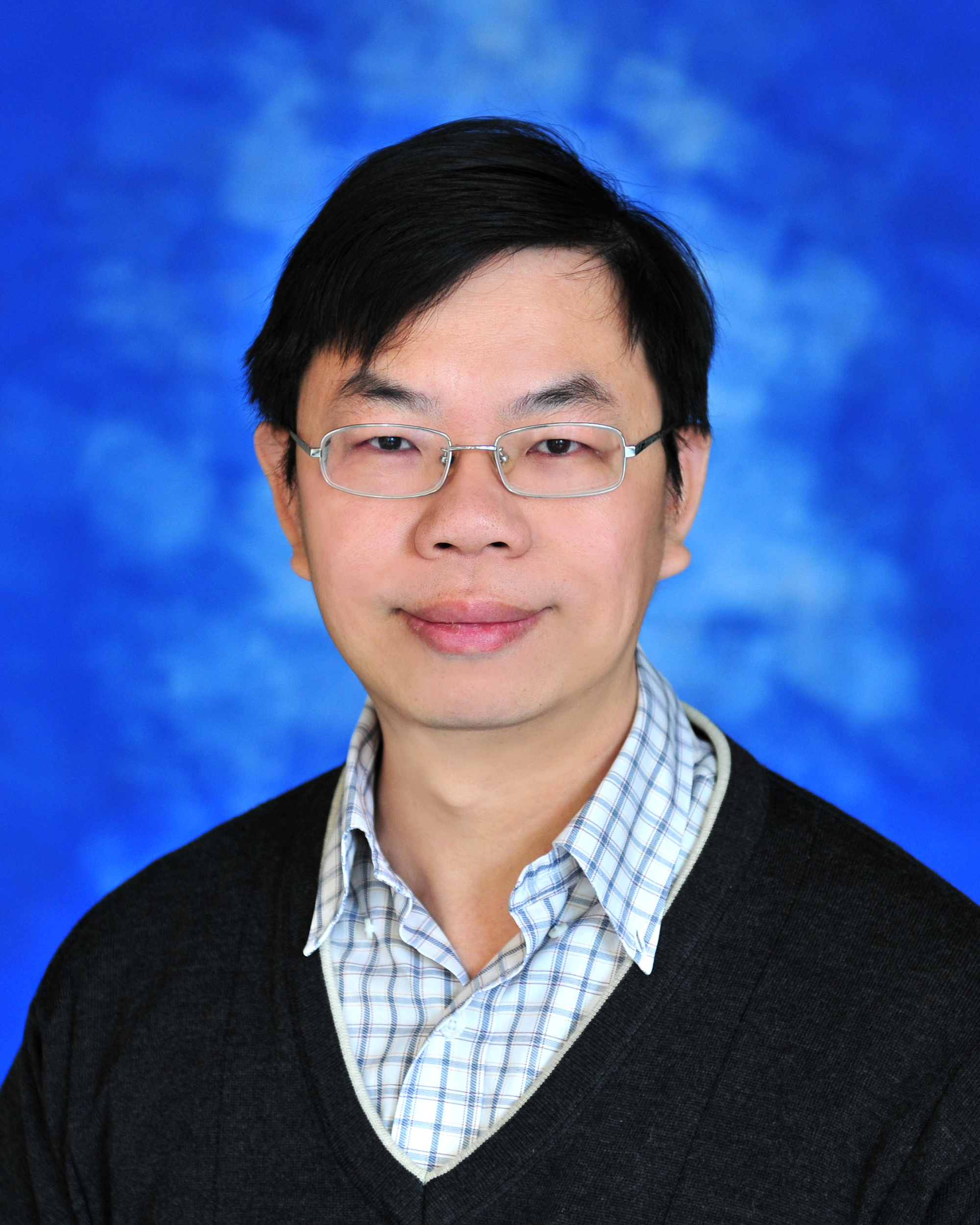}}]
	{Chia-Wen Lin}
	(Fellow, IEEE) received his Ph.D. degree in Electrical Engineering from National Tsing Hua University (NTHU), Hsinchu, Taiwan, in 2000.  
	Dr. Lin is currently a Professor with the Department of Electrical Engineering and the Institute of Communications Engineering, NTHU. He is also an R\&D Director with the Electronic and Optoelectronic System Research Laboratories, Industrial Technology Research Institute, Hsinchu, Taiwan. His research interests include image/video processing and computer vision.  He has served as Fellow Evaluating Committee member (2021--2022), BoG Member-at-Large (2022--2024), and Distinguished Lecturer (2018--2019) of IEEE Circuits and Systems society.   He was Chair of IEEE ICME Steering Committee (2020--2021). He served as TPC Co-Chair of IEEE ICIP 2019 and IEEE ICME 2010, and General Co-Chair of IEEE VCIP 2018.  He received two best paper awards from VCIP 2010 and 2015. He has served as an Associate Editor of \textsc{IEEE Transactions on Image Processing}, \textsc{IEEE Transactions on Circuits and Systems for Video Technology}, \textsc{IEEE Transactions on Multimedia}, and \textsc{IEEE Multimedia}. 
\end{IEEEbiography}

\end{document}